\documentclass[lettersize,journal]{IEEEtran}

\usepackage{amsmath,amsfonts}
\usepackage{algorithmic}
\usepackage{algorithm}
\usepackage{array}

\usepackage[caption=false,font=scriptsize ,labelfont=sf,textfont=sf]{subfig}

\usepackage{textcomp}
\usepackage{stfloats}
\usepackage{url}
\usepackage{verbatim}
\usepackage{graphicx}
\usepackage{cite}
\usepackage{booktabs}
\usepackage{hyperref}
\usepackage{amssymb}
\usepackage{multirow}
\usepackage{subcaption}
\usepackage{xcolor}
\usepackage{colortbl}
\usepackage{makecell}

\begin{document}

\title{ATSS: Detecting AI-Generated Videos via Anomalous Temporal Self-Similarity}


\author{Hang~Wang,~\IEEEmembership{Member,~IEEE,}
        Chao~Shen,~\IEEEmembership{Fellow,~IEEE,}
        Lei~Zhang,~\IEEEmembership{Fellow,~IEEE,} 
        and~Zhi-Qi~Cheng,~\IEEEmembership{Member,~IEEE}
\thanks{H. Wang is with Xi’an Jiaotong University, Xi’an, China, and The Hong Kong Polytechnic University, Hong Kong, China (e-mail: cshangwang@xjtu.edu.cn).}
\thanks{C. Shen is with Xi’an Jiaotong University, Xi’an, China (e-mail: chaoshen@mail.xjtu.edu.cn).}
\thanks{L. Zhang is with The Hong Kong Polytechnic University, Hong Kong, China (e-mail: cslzhang@comp.polyu.edu.hk).}
\thanks{Z.-Q. Cheng is with the School of Engineering and Technology, University of Washington, Tacoma, WA 98402, USA (e-mail: zhiqics@uw.edu).}
\thanks{Corresponding authors: Chao Shen; Zhi-Qi Cheng.}
}

\markboth{Journal of \LaTeX\ Class Files,~Vol.~14, No.~8, August~2021}%
{Shell \MakeLowercase{\textit{et al.}}: A Sample Article Using IEEEtran.cls for IEEE Journals}

\maketitle

\begin{abstract}
    AI-generated videos (AIGVs) have achieved unprecedented photorealism, posing severe threats to digital forensics.
    Existing AIGV detectors focus mainly on localized artifacts or short-term temporal inconsistencies, thus often fail to capture the underlying generative logic governing global temporal evolution, limiting AIGV detection performance.
    In this paper, we identify a distinctive fingerprint in AIGVs, termed anomalous temporal self-similarity (ATSS).
    Unlike real videos that exhibit stochastic natural dynamics, AIGVs follow deterministic anchor-driven trajectories (e.g., text or image prompts), inducing unnaturally repetitive correlations across visual and semantic domains.
    To exploit this, we propose the ATSS method, a multimodal detection framework that exploits this insight via a triple-similarity representation and a cross-attentive fusion mechanism.
    Specifically, ATSS reconstructs semantic trajectories by leveraging frame-wise descriptions to construct visual, textual, and cross-modal similarity matrices, which jointly quantify the inherent temporal anomalies.
    These matrices are encoded by dedicated Transformer encoders and integrated via a bidirectional cross-attentive fusion module to effectively model intra- and inter-modal dynamics.
    Extensive experiments on four large-scale benchmarks, including GenVideo, EvalCrafter, VideoPhy, and VidProM, demonstrate that ATSS significantly outperforms state-of-the-art methods in terms of AP, AUC, and ACC metrics, exhibiting superior generalization across diverse video generation models. Code and models of ATSS will be released. \footnote{~\url{https://github.com/hwang-cs-ime/ATSS}}
\end{abstract}


\section{Introduction}
\IEEEPARstart{R}{ecent} evolutions~\cite{xing2024survey, wang2025survey, verdoliva2020media} in video generation models, epitomized by text-to-video and image-to-video frameworks such as Runway Gen-4 \cite{runwayGen42025}, Show-1 \cite{zhang2025show}, Sora \cite{brooks2024video}, Stable Video Diffusion \cite{blattmann2023stable}, and Pika \cite{pika2022}, have significantly advanced the field of video synthesis, delivering unprecedented fidelity and temporal coherence.  
These foundation models enable users to generate high-resolution, cinematic-quality content, thereby revolutionizing media and creative industries such as advertising and entertainment \cite{chen2024videocrafter2, xing2024dynamicrafter, shi2024motion, li2025image}.
However, the widespread accessibility of video generation tools poses serious risks to the digital ecosystem, including the proliferation of disinformation~\cite{bao2018towards, guo2024pulid, zheng2023out, ijcai2021p157, zhao2023diffswap, oorloff2024avff, li2023learning}, the erosion of public trust~\cite{barrett2023identifying, sharma2023survey}, and intellectual property infringement.
As authentic and AI-generated videos (AIGVs) are becoming increasingly indistinguishable, developing generalizable detection frameworks has become  imperative for safeguarding information integrity.
Consequently, the task of AIGV detection has become an urgent necessity, aiming to defend downstream applications, such as news verification, forensics, and social media governance, against synthetic media threats.

\begin{figure}
\centering
\includegraphics[width=\linewidth]{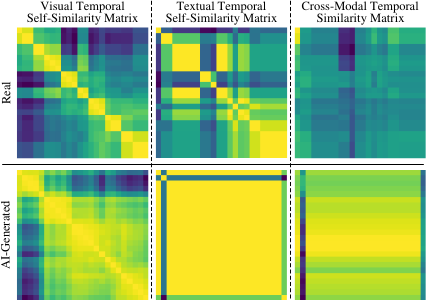} 
\caption{Motivation of the proposed ATSS framework. Real videos are characterized by stochastic spatiotemporal dynamics, resulting in diffuse and low-magnitude self-similarity patterns. In contrast, AI-generated videos exhibit systemic temporal regularity caused by their anchor-driven generation, which yields denser and higher-intensity correlation matrices across multiple modalities. This anomalous self-similarity serves as a distinctive forensic fingerprint, enabling ATSS to effectively distinguish AIGVs from natural sequences.}
\label{fig1_wh}
\end{figure}

AIGV detection remains fundamentally difficult due to two intrinsic challenges:
1) \textit{Open-Set Semantic Diversity.} 
Unlike traditional deepfakes that are constrained primarily to facial regions, AIGVs encompass arbitrary scenes, objects, and motions. 
This unconstrained semantic space renders fixed structural priors (e.g., facial landmarks) obsolete, necessitating a detection paradigm capable of generalizing across diverse visual contexts.
2) \textit{High-fidelity Temporal Coherence.} 
While prevailing detectors capitalize on sporadic flickers and fragmented inconsistencies, AIGVs exhibit systemic temporal coherence with remarkable photorealism.
This coherence motivates a transition from detecting localized artifacts to modeling the underlying generative logic that governs global evolution.

Existing synthetic video detection frameworks have explored various forensic traces, ranging from facial-centric manipulations to general AI-generated content. 
Specifically, traditional deepfake detectors~\cite{10.1145/3474085.3475508, zheng2021exploring, xu2023tall} utilize spatiotemporal modeling to capture inconsistencies, while recent AIGV-specific methods~\cite{bai2024ai, chen2024demamba, Zheng_2025_ICCV} shift toward capturing low-level flickering or physical law violations.
However, these approaches suffer from two critical bottlenecks that arise from the aforementioned challenges. 
First, they are predominantly content-agnostic and thus fail to anchor visual features to semantic trajectories. 
Whether constrained to facial priors or applied to general scenes, these detectors lack a semantic reference to distinguish natural evolution from generative anomalies in diverse open-set scenarios (\textit{Challenge 1}). 
Second, existing methods primarily target stochastic violations, such as localized artifacts or physical law inconsistencies, while ignoring the structured self-similarity inherent in high-fidelity video synthesis (\textit{Challenge 2}). 

To overcome these limitations, we shift the detection focus from discovering sporadic violations to modeling global generative patterns. 
We propose ATSS (anomalous temporal self-similarity), a novel framework that captures the generative logic of AIGVs through a triple-similarity representation and cross-attentive fusion scheme.
Our motivation originates from a key observation illustrated in Fig.~\ref{fig1_wh}: while authentic videos exhibit spontaneous and heterogeneous dynamics, AI-generated sequences follow deterministic anchor-driven trajectories.
This underlying generative process induces a distinctive fingerprint, which we term anomalous temporal self-similarity, manifested as dense and structured correlation patterns within and across visual and semantic domains.
Based on this observation, we formalize these temporal anomalies into a multimodal representational framework, i.e., ATSS.
Specifically, we first leverage a pretrained image captioning model to generate frame-wise descriptions, thereby reconstructing the semantic trajectories that govern the video evolution.
Subsequently, we construct three types of inter-frame similarity matrices, including visual, textual, and cross-modal, to quantify the systemic alignment between the synthesized scene and its semantic anchors.
The visual and textual matrices characterize temporal variations at the pixel and semantic levels, respectively, while the cross-modal matrix models the alignment between visual content and its reconstructed textual descriptions. 
Together, they expose intrinsic artifacts derived from forced generative coupling. 
To model multimodal temporal dynamics, these matrices are then encoded by dedicated Transformer encoders and integrated via a bidirectional cross-attentive fusion mechanism.
Finally, the resulting unified representation is fed into an MLP to predict the authenticity of the video.

To conclude, our contributions are summarized as follows:
\begin{itemize}
    \item We introduce anomalous temporal self-similarity as a distinctive forensic fingerprint for AIGV detection. We observe that AIGVs exhibit unnaturally structured and repetitive correlations within and across visual and semantic domains, a phenomenon that we attribute to the implicit constraints imposed by guiding anchors during the video synthesis process.

    \item We propose ATSS, a multimodal AIGV detection framework that models anchor-driven temporal evolution via a triple-similarity representation. By reconstructing semantic trajectories via frame-wise captioning, ATSS explicitly quantifies generative anomalies through visual, textual, and cross-modal inter-frame matrices. 
    These representations are then integrated via a cross-attentive fusion module to capture nuanced spatiotemporal dynamics.

    \item We conduct comprehensive experiments across four large-scale datasets, including GenVideo, EvalCrafter, VideoPhy, and VidProM. 
    Experimental results suggest that ATSS consistently outperforms existing state-of-the-art methods in AP, AUC, and ACC metrics, while maintaining superior generalization across diverse and unseen video generation models.
\end{itemize}

\begin{figure*}[!t]
\centering
\includegraphics[width=\textwidth]{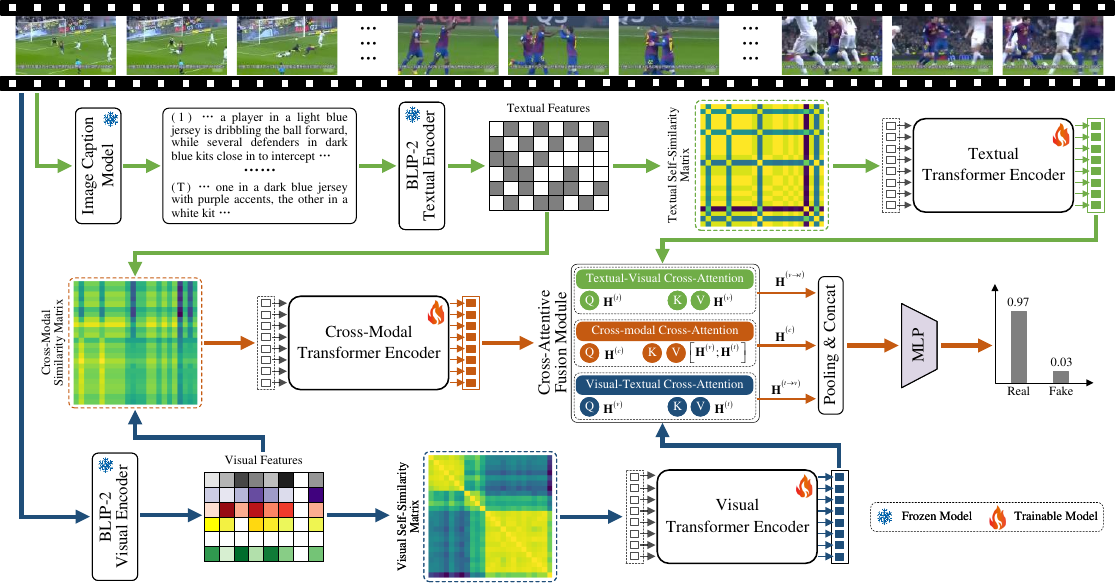} 
\caption{
The overall framework of ATSS. Given a video with \( T \) sampled frames, an image captioning model is first employed to generate frame-wise textual descriptions. Visual and textual features are then extracted to construct visual, textual, and cross-modal self-similarity matrices. Each matrix is processed by a dedicated Transformer to capture temporal dynamics. Finally, a cross-attentive fusion module integrates these modality-specific cues into a unified representation for binary classification.
}
\label{fig:pipeline}
\end{figure*}

\section{Related Work}
\label{sec:relat}

\subsection{Deepfake Video Detection}
Deepfake video detection targets identifying videos manipulated via facial forgery techniques, such as face-swapping and face-reenactment, by capturing localized biometric anomalies and spatiotemporal inconsistencies.
Existing methods have evolved from spatial artifact analysis to sophisticated spatiotemporal modeling. 
Specifically, MRE-Net~\cite{pang2023mre} and ISTVT~\cite{zhao2023istvt} utilize multi-rate excitation and interpretable Transformers to capture multi-scale temporal fluctuations. 
To enhance generalization, AltFreezing~\cite{wang2023altfreezing} alternately updates spatial and temporal weights to balance artifact extraction, while FakeSTormer~\cite{nguyen2025vulnerability} adopts vulnerability-aware learning to focus on subtle forgeries. 
For compressed or low-quality content, researchers have explored 3D spatiotemporal trajectories~\cite{chen2024compressed}, pixel-wise temporal frequency (BSF)~\cite{Kim_2025_ICCV}, and directional inconsistency patterns (DIP)~\cite{nie2024dip}. 
Moreover, FTCN~\cite{zheng2021exploring} and the work by Chen et al.~\cite{chen2022deepfake} emphasize temporal coherence by restricting spatial receptive fields and integrating attention mechanisms, respectively.
Yan et al.~\cite{yan2025generalizing} further introduced video-level blending to simulate facial feature drift, ensuring more robust generalized detection.

Beyond unimodal visual analysis, recent frameworks leverage multimodal cues and self-supervised paradigms to achieve universal detection. ID-Reveal~\cite{cozzolino2021id} and NACO~\cite{zhang2024learning} learn identity-specific motion or natural consistency representations from real videos to detect deviations in forged sequences. 
In the multimodal domain, AVFF~\cite{oorloff2024avff} and the self-supervised method by Feng et al.~\cite{feng2023self} capture audio-visual dissonance, such as lip-sync errors. 
To handle diverse scenes, UNITE~\cite{kundu2025towards} utilizes foundation models to unify detection across face and background manipulations, while MINTIME~\cite{10547206} addresses multi-identity scenarios through size-invariant embeddings.
Furthermore, Choi et al.~\cite{choi2024exploiting} exploited style latent flows to identify abnormal temporal shifts in facial attributes, and SFake~\cite{xie2024shaking} employs active probes to trigger mechanical vibrations for real-time verification.

Our ATSS differs from these deepfake detection methods in three aspects. 
First, while existing spatial-temporal methods target sporadic consistency violations or localized flaws (e.g., flickering), ATSS models systemic anomalous temporal self-similarity as a global fingerprint of anchor-driven synthesis. 
Second, unlike multimodal methods relying on simple audio-visual synchronization, ATSS constructs a triple-similarity representational space of visual, textual, and cross-modal modalities to explicitly quantify the deterministic semantic trajectories of guiding anchors.
Third, ATSS provides a content-agnostic framework that transcends facial-centric landmarks, effectively bridging the gap between traditional facial forensics and the challenging AIGV detection.

\subsection{AI-generated Video Detection}
AI-generated video (AIGV) detection aims to distinguish authentic videos from those synthesized videos using generative models, such as Diffusion models or GANs. 
Unlike traditional deepfake video detection which primarily focuses on specific biometric manipulations, AIGV detection seeks to identify forensic traces inherent in the generative process. 
Previous works \cite{bai2024ai, ji2024distinguish, 11210049, liu2024turns, chen2024demamba} accomplish  this task by capturing low-level motion inconsistencies and spatiotemporal artifacts.
Specifically, AIGVDet \cite{bai2024ai} employs a dual-branch architecture to learn anomalies in spatial and optical flow domains. Similarly, DuB3D \cite{ji2024distinguish} processes raw spatiotemporal sequences and GMFlow-based motion features to capture global inconsistencies.
To address the lack of temporal coherence by decoupling spatial artifacts, DeCoF \cite{11210049} utilizes pre-trained CLIP encoders to quantify frame-wise consistency and capture temporal anomalies.
Regarding diffusion-based video generators, DIVID \cite{liu2024turns} identifies synthesis artifacts by fusing diffusion reconstruction errors (DIRE) with spatiotemporal features through a CNN+LSTM architecture.
Additionally, DeMamba \cite{chen2024demamba} introduces a Detail Mamba module to efficiently capture spatial-temporal local inconsistencies, validated on the GenVideo benchmark.

Recent advancements \cite{zhang2025NSGVD, interno2025aigenerated, Zheng_2025_ICCV} have shifted toward modeling the intrinsic physical and geometric properties of video representations to improve detection generalization. 
NSG-VD \cite{zhang2025NSGVD} employs a physics-driven paradigm based on probability flow conservation, using Normalized Spatiotemporal Gradients (NSG) to detect violations of natural physical laws. 
From a geometric perspective, ReStraV \cite{interno2025aigenerated} exploits the ``perceptual straightening'' hypothesis by conducting a statistical analysis of trajectory curvature and step distance in the DINOv2 representation domain  to distinguish synthetic videos.
To reduce the dependency on extensive training data, D3 \cite{Zheng_2025_ICCV} introduces a training-free framework that performs second-order dynamical analysis inspired by Newtonian mechanics, identifying fundamental divergences in temporal feature distributions between real and synthetic videos.

Despite these developments, existing methods often treat generative artifacts as sporadic failures rather than systemic consequences of the underlying anchor-driven synthesis logic.
Our ATSS differs from these works in two aspects.
First, while most detectors focus on identifying fragmented artifacts or consistency violations, ATSS is the first to explicitly operationalize anomalous temporal self-similarity as a discriminative fingerprint, capturing the ``forced alignment'' inherent in anchor-driven generation. 
Second, unlike unimodal (e.g., visual-only) frameworks, ATSS reconstructs semantic trajectories of guiding anchors through frame-wise captioning. 
By constructing visual, textual, and cross-modal inter-frame similarity matrices, ATSS explicitly quantifies the deterministic evolution of semantic alignment. This multimodal representation provides a more comprehensive and interpretable forensic analysis, effectively bridging the gap between local artifact detection and global generative pattern modeling.

\section{Methodology}
\label{sec:method}
In this section, we elaborate on the proposed ATSS framework, which is designed to determine whether an input video is AI-generated or real.
Given a video with \( T \) uniformly sampled frames, denoted as $\mathcal{V} = \{I_1, I_2, \dots, I_T\}$, ATSS first applies an image captioning model BLIP-2 to each frame \( I_i \), yielding a sequence of frame-level captions.
Subsequently, we process these frames and frame-level captions using visual and textual encoders, respectively, and generate their corresponding embeddings.
Then, these embeddings are utilized to construct visual, textual, and cross-modal self-similarity matrices, which are fed into three dedicated Transformer encoders to extract spatiotemporal features within and across modalities.
Finally, a cross-attentive fusion mechanism operates on three-branch features to obtain the joint representations, which are fed into an MLP to predict the authenticity of given videos.
Fig.~\ref{fig:pipeline} illustrates the overall architecture of ATSS.

\subsection{Triplet Self-Similarity Matrices Construction}
Given a video, we first extract its corresponding visual and semantic features, and then construct visual, textual, and cross-modal self-similarity matrices to capture anomalous temporal self-similarity artifacts.

\noindent\textbf{Visual Representation.} 
For each frame \( I_t \), we use a visual encoder \( f_v(\cdot) \) based on the BLIP-2 vision model~\cite{li2023blip} to extract its visual representation, which is formulated as:
\begin{equation}
\mathbf{v}_t = f_v(I_t), \quad \mathbf{v}_t \in \mathbb{R}^{d}
\end{equation}

\noindent\textbf{Caption Generation \& Textual Representation.}
Regarding \( I_t \), we obtain its textual representation in two steps. First, we apply the image caption model \( g_c(\cdot) \) to generate a descriptive caption \( c_t \), which encapsulates the detailed semantics of the scene and objects:
\begin{equation}
c_t = g_c(I_t)
\end{equation}

Secondly, we utilize the textual encoder \( f_t(\cdot) \) to produce the textual embedding \( \mathbf{e}_t \) based on the frame-level caption \( c_t \):
\begin{equation}
\mathbf{e}_t = f_t(c_t), \quad \mathbf{e}_t \in \mathbb{R}^{d}
\end{equation}

Following the above process, we obtain visual and textual representations for each frame, which are temporally aligned.
These features are then used to construct three temporal self-similarity matrices, aimed at capturing temporal artifacts in the visual, textual semantic, and cross-modal domains.

\noindent\textbf{Triplet Self-Similarity Matrices Construction.} 
The extracted visual and textual representations of the video $ \mathcal{V} $, denoted as \( \{\mathbf{v}_t\}_{t=1}^{T} \) and \( \{\mathbf{e}_t\}_{t=1}^{T} \), are utilized to construct three types of self-similarity matrices: visual, textual, and cross-modal.

\begin{itemize}
    \item \textbf{Visual Self-similarity Matrix} \( \mathbf{S}^{(v)} \in \mathbb{R}^{T \times T} \): 
    We construct the visual similarity matrix by computing the cosine similarity of visual representations across all frames. 
    Each element \(\mathbf{S}^{(v)}_{i,j}\) measures the visual similarity between frame \(i\) and \(j\), which is defined as follows:
    \begin{equation}
    \mathbf{S}^{(v)}_{i,j} = \frac{\mathbf{v}_i^\top \mathbf{v}_j}{\|\mathbf{v}_i\|_2 \|\mathbf{v}_j\|_2}
    \end{equation}
    The visual similarity matrix is symmetric, with higher values indicating greater visual similarity between frame pairs. 
    In real videos, dynamic changes in scenes and objects, as well as shifts in camera perspective, result in significant similarity variations across frames. 
    In contrast, AI-generated videos are often constrained by text prompts or initial frames, leading to excessive temporal consistency and frame redundancy in visual dynamics. This results in an abnormally uniform distribution of high values within the visual self-similarity matrix \( \mathbf{S}^{(v)} \).
    Consequently, this matrix provides an effective representation to differentiate synthetic videos from real ones.

    \item \textbf{Textual Self-Similarity Matrix} \( \mathbf{S}^{(t)} \in \mathbb{R}^{T \times T} \): 
    Similarly, we generate the textual self-similarity matrix \( \mathbf{S}^{(t)} \) by calculating the cosine similarity between the textual features of every pair of frames. Each element \( \mathbf{S}^{(t)}_{i,j} \) denotes the semantic similarity between the $i$-th and $j$-th frames, defined as:
    \begin{equation}
    \mathbf{S}^{(t)}_{i,j} = \frac{\mathbf{e}_i^\top \mathbf{e}_j}{\|\mathbf{e}_i\|_2 \|\mathbf{e}_j\|_2}
    \end{equation}
    Since captions encode high-level semantics such as objects, actions, and scene context, \( \mathbf{S}^{(t)} \) captures temporal dynamics at the textual level. 
    Real videos typically involve evolving scenes and actions, leading to diverse captions and lower inter-frame textual similarity. 
    Conversely, AI-generated videos often exhibit similar frame-level captions due to consistent visual patterns, producing anomalously high and uniform values in \( \mathbf{S}^{(t)} \). 
    Thus, this matrix helps identify abnormal temporal-semantic patterns indicative of synthetic videos.

    \item \textbf{Cross-Modal Self-Similarity Matrix} \( \mathbf{S}^{(c)} \in \mathbb{R}^{T \times T} \): 
    This matrix models the semantic alignment between visual and textual modalities. 
    Each entry \( \mathbf{S}^{(c)}_{i,j} \) represents the cross-modal correlation between the visual embedding \( \mathbf{v}_i \) and the textual embedding \( \mathbf{e}_j \), computed by:
    \begin{equation}
    \mathbf{S}^{(c)}_{i,j} = \frac{\mathbf{v}_i^\top \mathbf{e}_j}{\|\mathbf{v}_i\|_2 \|\mathbf{e}_j\|_2}
    \end{equation}
    This matrix captures the semantic alignment between visual and textual features for every pair of frames.
    In real videos, visual and textual information often co-evolve over time, forming rich cross-modal patterns.
    However, AI-generated videos must adhere to given text prompts or initial frames during generation. 
    This constraint causes each frame's visual features to exhibit high similarity with captions from other frames, and vice versa.
    Therefore, \( \mathbf{S}^{(c)} \) provides complementary cross-modal clues to detect temporal artifacts in fake videos.
\end{itemize}

These self-similarity matrices encode fine-grained intra-modal and cross-modal spatial-temporal dynamics within videos. 
Intuitively, AI-generated videos tend to exhibit markedly repetitive similarity across frames, whereas real videos present more diverse inter-frame relationships. 
These matrices serve as input for the subsequent temporal modeling module, which is described next.

\subsection{Spatial-Temporal Modeling with Dedicated Transformer}
To further enhance temporal interactions within and across modalities, these three self-similarity matrices $\mathbf{S}^{(v)}$, $\mathbf{S}^{(t)}$, $\mathbf{S}^{(c)}$ are encoded by specific Transformer encoders.
Specifically, we consider each matrix as a sequence of $T$ tokens, where the $i$-th token corresponds to the $i$-th frame in video $\mathcal{V}$.
Formally, we define the token sequence as \(\mathbf{Z}^{(m)} = \mathbf{S}^{(m)} \in \mathbb{R}^{T \times T}\), where \(m \in \{v, t, c\}\).
Each sequence is independently processed by a dedicated Transformer to model intra-modal and inter-modal temporal dependencies and interactions, formulated as follows:
\begin{equation}
\mathbf{H}^{(m)} = \text{Transformer}^{(m)}(\mathbf{Z}^{(m)}), \quad m \in \{v, t, c\},
\end{equation}
where \(\mathbf{H}^{(m)} \in \mathbb{R}^{T \times d}\), $d=T$ is the channel dimension. 
\(\text{Transformer}^{(m)}\) is a modality-specific encoder composed of multiple layers of self-attention and feed-forward networks. 
These encoders operate directly on the tokenized self-similarity matrices, extracting higher-order temporal correlations and modality-specific patterns.

\subsection{Cross-Attentive Fusion Mechanism}
To aggregate the complementary contexts from the three-branch representations, we propose a bidirectional cross-attentive fusion mechanism. 
Specifically, it leverages textual features to refine the visual representations, and vice versa. 
The visual and textual contexts are incorporated in the cross-modal branch to mutually reinforce the representation.

First, we apply bidirectional cross-attention between the visual and textual streams to capture complementary artifacts:
\begin{equation}
\begin{aligned}
\tilde{\mathbf{H}}^{(t \rightarrow v)} &= \text{CrossAttn}(\mathbf{H}^{(v)}, \mathbf{H}^{(t)}, \mathbf{H}^{(t)}) \\
\tilde{\mathbf{H}}^{(v \rightarrow t)} &= \text{CrossAttn}(\mathbf{H}^{(t)}, \mathbf{H}^{(v)}, \mathbf{H}^{(v)})
\end{aligned}
\end{equation}
where \(\text{CrossAttn}(Q, K, V)\) denotes the standard multi-head attention operation with query \(Q\), key \(K\), and value \(V\), as illustrated in~\cite{vaswani2017attention}.

Subsequently, the cross-modal feature is refined by attending over the combined visual and textual embeddings: 
\begin{equation}
\tilde{\mathbf{H}}^{(c)} = \text{CrossAttn}(\mathbf{H}^{(c)}, [\mathbf{H}^{(v)};\mathbf{H}^{(t)}], [\mathbf{H}^{(v)};\mathbf{H}^{(t)}])
\label{eq:9}
\end{equation}
where $[\mathbf{H}^{(v)};\mathbf{H}^{(t)}] \in \mathbb{R}^{2T \times d}$ represents the concatenation of visual and textual features along the temporal dimension.
For instance, when cross-modal artifacts in AI-Generated videos appear deceptively diverse and resemble real video dynamics, Eq.~\ref{eq:9} amplifies the discriminative signal by attending to more pronounced artifact cues from the individual modalities.

The bidirectional cross-attentive fusion mechanism enables each branch to adaptively incorporate complementary cues from the others, thereby enhancing their individual discriminability. This strategy facilitates effective detection even when artifacts are asymmetrically distributed across modalities.

Then, we perform global average pooling over the temporal dimension for each enhanced representation:
\begin{equation}
\begin{aligned}
\mathbf{z}^{(t \rightarrow v)} &= \text{AvgPool}(\tilde{\mathbf{H}}^{(t \rightarrow v)}) \\
\mathbf{z}^{(v \rightarrow t)} &= \text{AvgPool}(\tilde{\mathbf{H}}^{(v \rightarrow t)}) \\
\mathbf{z}^{(c)} &= \text{AvgPool}(\tilde{\mathbf{H}}^{(c)})
\end{aligned}
\end{equation}

We obtain the final representation by concatenating these three pooled features:
\begin{equation}
\mathbf{z} = [\mathbf{z}^{(t \rightarrow v)}, \mathbf{z}^{(v \rightarrow t)}, \mathbf{z}^{(c)}] \in \mathbb{R}^{3d}
\end{equation}
where $[\cdot, \cdot]$ denotes concatenation along the channel dimension.

\subsection{Prediction Head and Loss}
Finally, the unified representation \( \mathbf{z} \) is passed through a multi-layer perceptron (MLP) followed by a softmax layer to obtain the final prediction:
\begin{equation}
\hat{\mathbf{y}}=\mathrm{Softmax}(\mathrm{MLP}(\mathbf{z}))=\big[\hat{y}_0,\hat{y}_1\big],
\end{equation}
where \(\hat{y}_0=P(\text{real}\mid\mathbf{z})\) and \(\hat{y}_1=P(\text{fake}\mid\mathbf{z})\).
We apply softmax to produce a normalized two-class probability distribution, satisfying \(\hat{y}_0+\hat{y}_1=1\). ATSS is trained with the standard cross-entropy loss:
\begin{equation}
\mathcal{L}
= -\,y\,\log\big(\hat{y}_1\big)\;-\;\big(1-y\big)\,\log\big(\hat{y}_0\big).
\end{equation}
where \(y\in\{0,1\}\) is the ground-truth label, with \(y=1\) denoting fake and \(y=0\) denoting real.

\begin{table*}[ht]
    \centering
    \caption{Comparison of detection AP ($\%$) between ATSS and 13 state-of-the-art baselines on GenVideo. \textsuperscript{\textdagger} Results are reproduced using the officially released code. \textsuperscript{\textdagger\textdagger} Results are reproduced from our implementation based on the original paper, as no official code is available.}
    \resizebox{\textwidth}{!}{
    \begin{tabular}{c || c c c c c c c c c c | c}
    \toprule
    \rowcolor{gray!25}
    Method                                                                            & MS        & MPS        & MV         & HotShot            & Show-1            & Gen2               & Crafter               & LaVie               & Sora                 & WS             & mean                \\
    \midrule
    \hline
    \rowcolor{gray!5}
    STIL\textsuperscript{\textdagger} \cite{10.1145/3474085.3475508}, ACM MM 2021               & 88.21             & 88.68              & 73.07              & 56.91              & 62.07             & 83.96              & 64.87                 & 65.54               & 49.86                & 63.60                  & 69.68               \\ 
    FTCN\textsuperscript{\textdagger} \cite{zheng2021exploring}, ICCV 2021            & 70.01             & 83.59              & 97.07              & 87.42              & 93.30             & 91.86              & 91.72                 & 84.16               & 44.48                & 84.46                  & 82.81               \\
    \rowcolor{gray!5} 
    X-CLIP\textsuperscript{\textdagger} \cite{ni2022expanding}, ECCV 2022             & 79.84             & 87.54              & 95.53              & 90.71              & 94.54             & 88.69              & 93.50                 & 86.28               & 64.23                & \underline{88.54}                  & 89.62               \\ 
    TALL\textsuperscript{\textdagger} \cite{xu2023tall}, ICCV 2023                    & 51.11             & 63.63              & 92.09              & 44.00              & 51.06             & 93.47              & 87.85                 & 59.07               & 15.82                & 64.43                  & 62.25               \\  
    \rowcolor{gray!5}
    FID\textsuperscript{\textdagger} \cite{NEURIPS2024_6dddcff5}, NeurIPS 2024        & 91.50 & 92.24              & 93.67              & 86.10  & 90.61 & 93.27              & 92.41                 & 83.68               & 74.95                & 82.24      & 88.07   \\ 
    NPR\textsuperscript{\textdagger} \cite{Tan_2024_CVPR}, CVPR 2024                  & 84.67             & 96.53              & 96.79              & 40.17              & 21.61             & 96.35              & 97.02                 & 22.37               & \underline{90.55}       & 66.51                  & 71.26               \\
    \rowcolor{gray!5}  
    MINTIME\textsuperscript{\textdagger} \cite{10547206}, TIFS 2024                  & 79.27             & 82.03              & 89.80              & 87.68              & 89.23             & 88.26              & 87.34                 & 82.48               & 80.75                & 85.10                  & 85.19               \\   
    AIGVDet\textsuperscript{\textdagger} \cite{bai2024ai}, PRCV 2024                  & 70.91             & 67.93              & 56.22              & 51.81              & 72.59             & 89.98              & 75.87                 & 88.62   & 65.70                & 64.96                  & 70.46               \\  
    \rowcolor{gray!5}
    DeMamba\textsuperscript{\textdagger} \cite{chen2024demamba}, arXiv 2024           & 41.96             & 97.07              & 84.64              & 67.63              & 45.07             & 96.11              & \underline{98.26}                 & 81.49               & 28.79                & 78.00                  & 71.90               \\
    DeCoF\textsuperscript{\textdagger\textdagger} \cite{11210049}, ICME 2025   & 91.18             & 91.69              & \underline{98.68}              & 76.02              & 48.99             & 98.28              & 94.67                 & 77.79               & 55.76                & 73.68                  & 81.67               \\
    \rowcolor{gray!5} 
    NSG-VD\textsuperscript{\textdagger} \cite{zhang2025NSGVD}, NeurIPS 2025        & 70.01 & 83.59 & 97.07 & 87.42 & 93.30 & 91.86 & 91.72 & 84.16 & 44.48 & 84.46 & 82.81  \\
    ReStraV\textsuperscript{\textdagger} \cite{interno2025aigenerated}, NeurIPS 2025            & \textbf{95.34}             & \underline{97.22}              & 97.43              & \underline{99.19}              & 74.52             & \textbf{99.92}             & 71.28             & \underline{96.80}             & \textbf{97.99}             & 73.75             & 90.34              \\
    \rowcolor{gray!5}  
    D3\textsuperscript{\textdagger} \cite{Zheng_2025_ICCV}, ICCV 2025        & 85.59             & 94.07             & 96.22             & 97.09             & \underline{95.29}             & 94.65             & 96.46             & 88.22             & 87.71             & 86.30             & \underline{92.16}         \\  
    \hline \hline
    \rowcolor{gray!25}
    \textbf{ATSS}                                             & \underline{95.21}     & \textbf{99.03}     & \textbf{99.68}              & \textbf{99.61}     & \textbf{99.72}    & \underline{99.68}     & \textbf{99.94}        & \textbf{98.51}      & 85.75    & \textbf{98.48}         & \textbf{97.56}      \\ 
    \bottomrule
    \end{tabular}
    }
\label{tab:ap_GenVideo}
\end{table*}

\begin{table*}[ht]
    \centering
    \caption{Comparison of detection AUC ($\%$) between ATSS and 13 state-of-the-art baselines on GenVideo. \textsuperscript{\textdagger} Results are reproduced using the officially released code. \textsuperscript{\textdagger\textdagger} Results are reproduced from our implementation based on the original paper, as no official code is available.}
    \resizebox{\textwidth}{!}{
    \begin{tabular}{c || c c c c c c c c c c | c}
    \toprule
    \rowcolor{gray!25}
    Method                                                         & MS     & MPS     & MV      & HotShot           & Show-1            & Gen2                 & Crafter                 & LaVie                 & Sora                 & WS                 & mean   \\
    \midrule 
    \hline
    \rowcolor{gray!5}
    STIL\textsuperscript{\textdagger} \cite{10.1145/3474085.3475508}, ACM MM 2021               & 86.48          & 88.75           & 82.77           & 60.00             & 68.71             & 86.67                & 73.56                   & 72.67               & 54.18                & 69.94                  & 74.37     \\
    FTCN\textsuperscript{\textdagger} \cite{zheng2021exploring}, ICCV 2021                      & 69.76          & 83.25           & 97.18           & 88.69             & 93.45             & 93.33                & 91.62                   & 84.70               & 38.30                & 85.74                  & 82.60     \\
    \rowcolor{gray!5}
    X-CLIP\textsuperscript{\textdagger} \cite{ni2022expanding}, ECCV 2022                       & 80.51          & 86.14           & 95.12           & 91.51             & \underline{94.69}             & 88.96                & 92.84                   & 85.04               & 64.09                & \underline{87.93}                  & 88.87     \\  
    TALL\textsuperscript{\textdagger} \cite{xu2023tall}, ICCV 2023                              & 58.46          & 60.54           & 83.24           & 45.21             & 46.24             & 73.30                & 66.37                   & 48.40               & 66.36                & 53.84                  & 60.20     \\  
    \rowcolor{gray!5}
    FID\textsuperscript{\textdagger} \cite{NEURIPS2024_6dddcff5}, NeurIPS 2024                  & 90.94          & 91.93           & 93.70           & 85.77       & 91.19       & 93.16                & 92.02                   & 82.57     & 73.45                & 81.30          & 87.60  \\ 
    NPR\textsuperscript{\textdagger} \cite{Tan_2024_CVPR}, CVPR 2024                            & 93.92 & \textbf{99.38}  & \textbf{99.92}  & 26.83             & 18.99             & 98.78                & \underline{99.24}       & 42.32                 & \textbf{97.56}       & 76.12                      & 75.31  \\
    \rowcolor{gray!5}
    MINTIME\textsuperscript{\textdagger} \cite{10547206}, TIFS 2024                            & 81.27             & 85.48              & 90.97              & 90.34              & 90.89             & 88.80              & 89.77                 & 83.86               & 79.27                & 85.71                  & 86.64               \\ 
    AIGVDet\textsuperscript{\textdagger} \cite{bai2024ai}, PRCV 2024                            & 68.05          & 79.42           & 59.41           & 74.67             & 70.29             & 71.62                & 69.77                   & 79.55                 & 60.79                & 67.82                      & 70.14  \\ 
    \rowcolor{gray!5} 
    DeMamba\textsuperscript{\textdagger} \cite{chen2024demamba}, arXiv 2024                     & 80.53          & 95.52           & \underline{99.90}  & 59.49          & 49.34             & 99.10    & 97.99                   & 63.25                 & 91.76    & 71.15                      & 80.80  \\ 
    DeCoF\textsuperscript{\textdagger\textdagger} \cite{11210049}, ICME 2025   & \underline{95.49}          & 96.87           & 99.60           & 90.22             & 77.86             & 98.94                & 96.63                   & 84.62                 & 96.36                & 87.72                      & 92.43   \\
    \rowcolor{gray!5}
    NSG-VD\textsuperscript{\textdagger} \cite{zhang2025NSGVD}, NeurIPS 2025      & 69.76 & 83.25 & 97.18 & 88.69 & 93.45 & 93.33 & 91.62 & 84.70 & 38.30 & 85.74 & 82.60  \\
    ReStraV\textsuperscript{\textdagger} \cite{interno2025aigenerated}, NeurIPS 2025            & \textbf{96.84}             & 98.20              & 96.95              & \underline{99.27}              & 79.65             & \textbf{99.93}             & 77.40             & \underline{97.75}             & \underline{97.35}             & 81.55             & \underline{92.49}              \\   
    \rowcolor{gray!5}
    D3\textsuperscript{\textdagger} \cite{Zheng_2025_ICCV}, ICCV 2025        & 86.00           & 93.62           & 95.74           & 96.96           & 94.03           & 93.93           & 96.21           & 88.93           & 88.93           & 87.60           & 92.20       \\
    \hline \hline
    \rowcolor{gray!25}
    \textbf{ATSS}                                             & 93.49 & \underline{98.82}  & 99.62      & \textbf{99.44}    & \textbf{99.67}    & \underline{99.58}       & \textbf{99.92}          & \textbf{97.95}        & 84.89                & \textbf{97.76}             & \textbf{97.11}  \\ 
    \bottomrule
    \end{tabular}
    }
\label{tab:auc_GenVideo}
\end{table*}

\section{Experiments}
\label{sec:exper}

\subsection{Datasets}
Four large-scale benchmark datasets GenVideo~\cite{chen2024demamba}, EvalCrafter~\cite{liu2024evalcrafter}, VideoPhy~\cite{bansal2024videophy}, and VidProM~\cite{wang2024vidprom} are used to evaluate the performance of our proposed ATSS approach.

\noindent\textbf{Training Set.}
The training set is obtained from the training split of the GenVideo dataset~\cite{chen2024demamba}, comprising 953,279 real videos randomly sampled from Youku-mPLUG~\cite{xu2023youku} and 98,377 AI-generated videos using the official Pika model~\cite{pika2022}.

\noindent\textbf{Validation Set.}
We randomly select 10\% of the training data as the validation set. 
It is used mainly for adaptive learning rate adjustment and hyperparameter tuning during training.

\noindent\textbf{Testing Set.}
To verify the generalization of ATSS to unseen generative models and scenarios, we conduct extensive evaluations on AI-generated videos drawn from 40 distinct subsets spanning four large-scale benchmarks: GenVideo, EvalCrafter, VideoPhy, and VidProM. 
These subsets are created by a wide variety of SOTA video generators, providing a challenging set of out-of-distribution (OOD) tests.
The detailed composition of each subset is described as follows:

i) \textit{GenVideo}~\cite{chen2024demamba}: encompasses fake videos produced by 10 video generators: MoonValley (MV) ~\cite{moonvalley2022}, ModelScope (MS)~\cite{wang2023modelscope}, MorphStudio (MPS)~\cite{morph2023}, Show-1~\cite{zhang2024show}, HotShot~\cite{hotshot2023}, Gen2~\cite{esser2023structure}, LaVie~\cite{wang2025lavie}, Sora~\cite{brooks2024video}, Crafter~\cite{chen2023videocrafter1}, and WildScrape (WS) ~\cite{wei2024dreamvideo, feng2023dreamoving, xu2024magicanimate}.

ii) \textit{EvalCrafter}~\cite{liu2024evalcrafter}: provides AI-Generated videos from 14 generators: VideoCrafter V0.9 (Floor33), MoonValley (MV), Gen2, Mix-SR, Gen2-December (Gen2-D), HotShot-XL (HS-XL), Show-1, ModelScope (MS), LaVie-Base (LaVie-B), LaVie-Interpolation (LaVie-I), PikaLab (PKL), VideoCrafter (VC), PikaLab V1.0 (PKL-V1), and ZeroScope (ZS).

iii) \textit{VideoPhy}~\cite{bansal2024videophy}: contains synthetic videos generated by 10 models: CogVideoX (CVX), Gen2, Dream-Machine (DM), VideoCrafter2 (VC2), LaVie, SVD-T2I2V, OpenSora, CogVideoX-5B (CVX-5B), Pika, and ZeroScope (ZS).

iv) \textit{VidProM}~\cite{wang2024vidprom}: comprises 6 fake subsets generated by Text2Video-Zero (T2VZ), Pika, ModelScope (MS), StreamingT2V (ST2V), VideoCrafter2 (VC2), and OpenSora.

Following the GenVideo testing setup, we adopt MSR-VTT~\cite{xu2016msr} as the source of real videos.

\begin{table*}[th]
    \centering
    \caption{Comparison of detection AP ($\%$) between ATSS and 13 state-of-the-art baselines on EvalCrafter.  \textsuperscript{\textdagger} Results are reproduced using the officially released code. \textsuperscript{\textdagger\textdagger} Results are reproduced from our implementation based on the original paper, as no official code is available.}
    \resizebox{\textwidth}{!}{
    \begin{tabular}{c || c c c c c c c c c c c c c c| c}
    \toprule
    \rowcolor{gray!25}
    Method &  \makecell[c]{\rotatebox{50}{Floor33}} & \makecell[c]{\rotatebox{50}{Gen2}} &  \makecell[c]{\rotatebox{50}{Gen2-D}} &  \makecell[c]{\rotatebox{50}{HS-XL}} &  \makecell[c]{\rotatebox{50}{LaVie-B}} &  \makecell[c]{\rotatebox{50}{LaVie-I}} &  \makecell[c]{\rotatebox{50}{Mix-SR}} &  \makecell[c]{\rotatebox{50}{MS}} &  \makecell[c]{\rotatebox{50}{MV}} &  \makecell[c]{\rotatebox{50}{PKL}} &  \makecell[c]{\rotatebox{50}{PKL-V1}} &  \makecell[c]{\rotatebox{50}{Show-1}} &  \makecell[c]{\rotatebox{50}{VC}} & \makecell[c]{\rotatebox{50}{ZS}} &  \makecell[c]{\rotatebox{50}{mean}} \\
    \midrule
    \hline
    \rowcolor{gray!5}
    STIL\textsuperscript{\textdagger}               & 90.42        & 90.59        & 68.07        & 59.36        & 65.96        & 66.02        & 65.41        & 89.20        & 74.71        & 95.86        & 79.97        & 60.21        & 67.14        & 89.73        & 75.90        \\
    FTCN\textsuperscript{\textdagger}               & 82.85        & 88.50        & 94.72        & 88.25        & 82.71        & 81.91        & 92.74        & 71.16        & 96.65        & 96.36        & 93.14        & 93.55        & 92.23        & 69.09        & 87.42        \\
    \rowcolor{gray!5}
    X-CLIP\textsuperscript{\textdagger}             & 87.32        & 87.56        & 90.72        & 89.67        & 85.58        & 83.55        & 94.22        & 79.45        & 96.04        & 97.44        & 96.15        & 95.68        & 91.42        & 77.63        & 90.28        \\   
    TALL\textsuperscript{\textdagger}               & 63.25        & 70.75        & 77.04        & 46.93        & 52.87        & 52.53        & 78.16        & 62.11        & 83.63        & 65.33        & 70.98        & 48.00        & 60.50        & 51.73        & 63.13        \\  
    \rowcolor{gray!5}
    FID\textsuperscript{\textdagger}                & 96.40        & 97.36        & 98.68        & 89.90        & 92.92        & 84.19        & 98.51        & 95.74        & 98.29        & 99.49        & 99.17        & \underline{96.77}        & 95.71        & 95.18        & \underline{95.59}        \\ 
    NPR\textsuperscript{\textdagger}                & \textbf{99.77}        & \underline{99.34}        & \textbf{99.95}        & 47.39        & 76.45        & 72.23        & \textbf{99.67}   & \textbf{98.54}        & \textbf{99.96}        & \underline{99.97}        & \textbf{99.93}        & 69.82        & \textbf{99.68}        & \underline{98.21}        & 90.07        \\ 
    \rowcolor{gray!5} 
    MINTIME\textsuperscript{\textdagger}           & 84.62        & 86.20        & 88.02        & 90.07        & 83.12        & 82.99        & 88.87        & 78.47        & 90.56        & 94.65        & 91.31        & 87.99         & 88.18        & 88.43        & 87.39        \\  
    AIGVDet\textsuperscript{\textdagger}            & 67.84        & 71.86        & 74.24        & 51.46        & 73.81        & 70.72        & 57.64        & 71.00        & 56.50        & 94.95        & 92.92        & 72.41        & 64.58        & 67.00        & 70.50        \\
    \rowcolor{gray!5}   
    DeMamba\textsuperscript{\textdagger}            & 97.50        & 89.82        & 97.67        & 66.31        & 75.37        & 69.51        & 96.38        & 41.13        & 85.79        & 70.08        & 34.32        & 46.13        & 98.19        & 96.34        & 76.04        \\
    DeCoF\textsuperscript{\textdagger\textdagger}   & 91.69             & 96.30              & 97.18              & 76.02              & 70.40             & 59.39              & 93.77                 & 91.18               & 98.68                & 98.70                  & 98.43                 & 48.99               & 86.75                & 83.85                  & 85.10               \\
    \rowcolor{gray!5}
    NSG-VD\textsuperscript{\textdagger}             & 82.85 & 88.50 & 94.72 & 88.25 & 82.71 & 81.91 & 92.74 & 71.16 & 96.65 & 96.36 & 93.14 & 93.55 & 92.23 & 69.09 & 87.42  \\
    ReStraV\textsuperscript{\textdagger}            & 97.72             & \textbf{99.93}              & \underline{99.94}              & \underline{99.21}              & \underline{95.70}             & \underline{97.35}             & 70.64             & \underline{96.04}             & 96.99             & \textbf{100.00}             & \underline{99.71}     & 77.14             & 75.18             & 63.45             & 90.64              \\ 
    \rowcolor{gray!5}  
    D3\textsuperscript{\textdagger}                 & 94.09           & 93.36           & 96.22           & 96.96           & 89.37           & 87.28           & 96.17           & 86.70           & 96.74           & 94.34           & 95.02           & 94.16           & 95.78           & 94.84           & 93.64       \\
    \hline \hline
    \rowcolor{gray!25}
    \textbf{ATSS}                               & \underline{99.09}        & 98.89        & 99.46        & \textbf{99.43}        & \textbf{97.28}        & \textbf{97.69}        & \underline{98.91}        & 93.60        & \underline{99.37}        & 99.80        & 99.34        & \textbf{99.68}        & \underline{99.56}        & \textbf{98.50}        & \textbf{98.61}        \\  
    \bottomrule
    \end{tabular}
    }
\label{tab:ap_EvalCrafter}
\end{table*}

\begin{table*}[th]
    \centering
    \caption{Comparison of detection AUC ($\%$) between ATSS and 13 state-of-the-art baselines on EvalCrafter.  \textsuperscript{\textdagger} Results are reproduced using the officially released code. \textsuperscript{\textdagger\textdagger} Results are reproduced from our implementation based on the original paper, as no official code is available.}
    \resizebox{\textwidth}{!}{
    \begin{tabular}{c || c c c c c c c c c c c c c c| c}
    \toprule
    \rowcolor{gray!25}
    Method &  \makecell[c]{\rotatebox{50}{Floor33}} & \makecell[c]{\rotatebox{50}{Gen2}} &  \makecell[c]{\rotatebox{50}{Gen2-D}} &  \makecell[c]{\rotatebox{50}{HS-XL}} &  \makecell[c]{\rotatebox{50}{LaVie-B}} &  \makecell[c]{\rotatebox{50}{LaVie-I}} &  \makecell[c]{\rotatebox{50}{Mix-SR}} &  \makecell[c]{\rotatebox{50}{MS}} &  \makecell[c]{\rotatebox{50}{MV}} &  \makecell[c]{\rotatebox{50}{PKL}} &  \makecell[c]{\rotatebox{50}{PKL-V1}} &  \makecell[c]{\rotatebox{50}{Show-1}} &  \makecell[c]{\rotatebox{50}{VC}} & \makecell[c]{\rotatebox{50}{ZS}} &  \makecell[c]{\rotatebox{50}{mean}} \\
    \midrule
    \hline
    \rowcolor{gray!5}
    STIL\textsuperscript{\textdagger}               & 90.14             & 91.40             & 79.25             & 61.62             & 73.63             & 71.23             & 74.22             & 87.28             & 84.22             & 97.82     & 88.50             & 66.68             & 74.49             & 90.77     & 80.80          \\ 
    FTCN\textsuperscript{\textdagger}               & 82.78             & 90.76             & 95.11             & 90.29             & 84.51             & 83.00             & 92.31             & 70.99             & 96.65             & 96.73     & 94.01             & 93.62             & 91.27             & 70.82      & 88.06          \\
    \rowcolor{gray!5}
    X-CLIP\textsuperscript{\textdagger}             & 86.10             & 86.70             & 90.93             & 90.80             & 84.06             & 83.01             & 93.59             & 79.69             & 95.56             & 97.45     & 95.79             & 95.29             & 90.71             & 80.53      & 89.38          \\   
    TALL\textsuperscript{\textdagger}               & 61.46             & 72.04             & 74.60             & 43.88             & 49.64             & 50.02             & 77.50             & 58.99             & 83.19             & 67.13             & 72.51             & 46.81             & 55.69             & 46.03             & 61.39             \\ 
    \rowcolor{gray!5}
    FID\textsuperscript{\textdagger}                & 96.26      & 97.46         & 98.58        & 90.05       & 92.32     & 83.42     & 98.23           & 95.27     & 98.28 & 99.48     & 99.24 & \underline{96.79} & 95.26 & \underline{95.29}     & \underline{95.42}     \\ 
    NPR\textsuperscript{\textdagger}                & \textbf{99.38}     & 97.74             & \underline{99.80}     & 26.83             & 50.23             & 36.00             & \underline{99.22}             & 93.92             & \textbf{99.92}     & \underline{99.92}     & \underline{99.71}     & 18.99             & \textbf{99.25}     & 92.60     & 79.54             \\ 
    \rowcolor{gray!5}
    MINTIME\textsuperscript{\textdagger}           & 86.88             & 88.01             & 89.41             & 91.71             & 84.07             & 84.23             & 90.16             & 79.60             & 91.68             & 95.08      & 92.33            & 90.08               & 90.16            & 91.29      & 88.91               \\ 
    AIGVDet\textsuperscript{\textdagger}            & 79.52             & 70.10             & 73.03             & 73.92             & 85.34 & 74.73 & 64.44             & 68.17             & 59.90             & 92.78             & 90.60             & 70.23             & 74.92             & 67.75             & 74.67             \\ 
    \rowcolor{gray!5}  
    DeMamba\textsuperscript{\textdagger}                      & 95.41             & 98.60     & 99.70     & 59.10             & 72.87             & 55.39             & 98.91     & 79.11             & \underline{99.87}     & 99.51 & 99.70     & 48.89             & 97.08             & 76.98             & 84.37             \\
    DeCoF\textsuperscript{\textdagger\textdagger}   & 96.87             & \underline{98.69}              & 99.17              & 90.22              & 87.71             & 81.53              & 98.05                 & \underline{95.49}               & 99.60                & 99.54                  & 99.42                 & 77.89               & 95.20                & 92.20                  & 93.68               \\
    \rowcolor{gray!5}
    NSG-VD\textsuperscript{\textdagger}             & 82.78 & 90.76 & 95.11 & 90.29 & 84.51 & 83.00 & 92.31 & 70.99 & 96.65 & 96.73 & 94.01 & 93.62 & 91.27 & 70.82 & 88.06  \\
    ReStraV\textsuperscript{\textdagger}            & 98.33             & \textbf{99.93}              & \textbf{99.94}              & \underline{99.13}              & \underline{96.93}             & \textbf{97.97}              & 73.98              & \textbf{97.43}              & 96.53             & \textbf{100.00}              & \textbf{99.85}              & 81.71              & 82.05             & 68.22             & 92.29              \\ 
    \rowcolor{gray!5}  
    D3\textsuperscript{\textdagger}                 & 93.68           & 92.24           & 95.85           & 96.93           & 90.06           & 88.68           & 95.62           & 87.38           & 96.32           & 93.77           & 94.73           & 92.49           & 95.58           & 93.76           & 93.36              \\
    \hline \hline
    \rowcolor{gray!25}
    \textbf{ATSS}                               & \underline{98.92} & 98.60     & 99.28 & \textbf{99.23}     & \textbf{96.99} & \underline{96.79} & \textbf{99.57}     & 91.83 & 99.08 & 99.79             & 99.61             & \textbf{99.63}     & \textbf{99.53}     & \textbf{98.02}             & \textbf{98.35}     \\  
    \bottomrule
    \end{tabular}
    }
\label{tab:auc_EvalCrafter}
\end{table*}

\subsection{Implementation Details}
The training of ATSS is conducted on a single NVIDIA L40S GPU for a total of 200 epochs.
The number of frames uniformly sampled from each video is set to $T=8$.
For frame-level caption generation, we employ the pre-trained BLIP-2 model with the ``blip2-flan-t5-xl'' weights.
For the extraction of visual and textual features, we initialize the BLIP-2 visual encoder and BLIP-2 textual encoder using the ``blip2-itm-vit-g'' weights. These encoders remain frozen throughout the training phase.

Three independent Transformer encoders are employed to process different types of self-similarity matrices. Each encoder consists of 2 layers with 4 attention heads, and both the hidden and feed-forward dimensions are set to 32. 
All weights are initialized using the Xavier uniform distribution while all biases are set to zero. 
These encoders are trained from scratch without any parameter sharing.

The model is optimized with the Adam~\cite{2015-kingma} algorithm with an initial learning rate of $1 \times 10^{-4}$, which is dynamically adjusted using a ``ReduceLROnPlateau'' scheduler in `max' mode, with a reduction factor of 0.5 and patience of 3 epochs.

\subsection{Baselines}
We compare the proposed ATSS with 13 baselines, comprising 2 image-level and 11 video-level detectors. 
Specifically, the latter are further categorized into 5 deepfake video detection methods (STIL, FTCN, X-CLIP, TALL, MINTIME) and 6 AI-generated video detection frameworks (AIGVDet, DeMamba, DeCoF, NSG-VD, ReStraV, and D3). A brief overview of each baseline is provided as follows:

\begin{itemize}
    \item STIL \cite{10.1145/3474085.3475508}: learns spatiotemporal inconsistencies between consecutive frames to identify deepfake videos.
    
    \item FTCN \cite{zheng2021exploring}: captures short-range temporal inconsistencies using local attention over consecutive frames for generalized face forgery detection.
    
    \item X-CLIP \cite{ni2022expanding}: extends CLIP with temporal modeling to align frame-wise visual and textual features for general deepfake video detection.
    
    \item TALL \cite{xu2023tall}: arranges multiple frames into a thumbnail layout and processes them jointly to detect deepfakes via spatial-temporal representation learning.

    \item FID \cite{NEURIPS2024_6dddcff5}: utilizes features extracted from Inception-V3 \cite{szegedy2016rethinking} to distinguish real and fake images, and aggregates frame-level results for video-level prediction.
    
    \item NPR \cite{Tan_2024_CVPR}: identifies CNN-specific up-sampling artifacts to recognize AI-generated images, with video-level predictions obtained via temporal aggregation.
    
    \item MINTIME \cite{10547206}: captures temporal inconsistencies across diverse facial subjects with a size-invariant design for robust deepfake detection.

    \item AIGVDet \cite{bai2024ai}: detects AI-generated videos by jointly modeling spatial anomalies in RGB frames and temporal inconsistencies in optical flow via a two-branch network with decision-level fusion.
    
    \item DeMamba \cite{chen2024demamba}: employs a Mamba-based architecture to model long-range frame-wise dependencies for effective AI-generated video detection.
    
    \item DeCoF \cite{11210049}: utilizes frame-pair similarity to model temporal consistency across frames for detecting AI-generated videos.

    \item NSG-VD \cite{zhang2025NSGVD}: proposes a Normalized Spatiotemporal Gradient (NSG) statistic and employs Maximum Mean Discrepancy (MMD) on NSG features to detect AI-generated videos.

    \item ReStraV \cite{interno2025aigenerated}: leverages the perceptual straightening hypothesis to detect AI-generated videos by quantifying the temporal curvature of video trajectories in a latent representation space.

    \item D3 \cite{Zheng_2025_ICCV}: adopts a training-free approach that detects AI-generated videos by computing the standard deviation of second-order temporal visual features.
\end{itemize}

For a fair comparison, all baseline methods and our proposed ATSS are evaluated using the identical dataset partitioning protocol.

\begin{table*}[!ht]
    \centering
    \caption{Comparison of detection AP ($\%$) between ATSS and 13 state-of-the-art baselines on VideoPhy.  \textsuperscript{\textdagger} Results are reproduced using the officially released code. \textsuperscript{\textdagger\textdagger} Results are reproduced from our implementation based on the original paper, as no official code is available.}
    \resizebox{\textwidth}{!}{
    \begin{tabular}{c || c c c c c c c c c c| c}
    \toprule
    \rowcolor{gray!25}
    Method                                                                                                 & CVX   & CVX-5B   & DM   & Gen-2        & LaVie        & OpenSora     & Pika       & SVD-T2I2V    & VC2     & ZS    & mean   \\
    \midrule
    \hline
    \rowcolor{gray!5}
    STIL\textsuperscript{\textdagger} \cite{10.1145/3474085.3475508}, ACM MM 2021                          & 66.43        & 69.61         & 65.31           & 69.63        & 63.68        & 61.94        & 97.07      & 67.51        & 67.54             & 58.56        & 68.73   \\
    FTCN\textsuperscript{\textdagger} \cite{zheng2021exploring}, ICCV 2021                                 & 74.24        & 74.83         & 67.98           & 93.99        & 69.16        & 66.95        & 93.94      & 82.81        & 83.39             & 75.64        & 78.29   \\
    \rowcolor{gray!5}
    X-CLIP\textsuperscript{\textdagger} \cite{ni2022expanding}, ECCV 2022                                  & 87.35        & 83.72         & 75.54           & 92.09        & 79.52        & 90.44        & 96.13      & 85.26        & 85.35             & 88.88        & 86.66   \\   
    TALL\textsuperscript{\textdagger} \cite{xu2023tall}, ICCV 2023                                         & 39.59        & 50.72         & 62.36           & 70.78        & 40.40        & 37.30        & 62.69      & 52.62        & 52.66             & 50.66        & 51.98   \\
    \rowcolor{gray!5}
    FID\textsuperscript{\textdagger} \cite{NEURIPS2024_6dddcff5}, NeurIPS 2024                             & \textbf{93.34}      & 91.41         & 97.50   & 98.35         & \underline{96.51}    & 87.90 & 99.55 & 95.66 & 96.03     & 90.60 & \underline{94.69} \\
    NPR\textsuperscript{\textdagger} \cite{Tan_2024_CVPR}, CVPR 2024                                       & 81.37        & 81.99         & \textbf{99.86}   & \underline{99.90} & 63.72        & 88.78    & \underline{99.91} & \textbf{99.54}        & 60.21             & 78.23        & 85.35   \\
    \rowcolor{gray!5}
    MINTIME\textsuperscript{\textdagger} \cite{10547206}, TIFS 2024                                       & 85.27             & 85.96              & 77.60              & 84.12              & 79.16             & 90.64              & 92.86                 & 79.01               & 83.83                & 86.34                  & 84.48               \\
    AIGVDet\textsuperscript{\textdagger} \cite{bai2024ai}, PRCV 2024                                       & 63.15        & 58.95         & 59.27           & 61.55        & 61.06        & 59.07        & 92.96      & 53.73        & 58.22             & 63.11        & 63.11   \\
    \rowcolor{gray!5}
    DeMamba\textsuperscript{\textdagger} \cite{chen2024demamba}, arXiv 2024                                & 22.10        & 15.96         & 72.52           & 92.92        & 50.00        & \underline{94.70}        & 45.47      & 91.59        & 90.15             & 73.64        & 64.91   \\
    DeCoF\textsuperscript{\textdagger\textdagger} \cite{11210049}, ICME 2025                               & 14.35             & 17.10              & 79.20              & 88.81              & 42.72             & 19.92              & 95.01                 & 59.68               & 46.13                & 21.59                  & 48.45               \\
    \rowcolor{gray!5}
    NSG-VD\textsuperscript{\textdagger} \cite{zhang2025NSGVD}, NeurIPS 2025               & 74.24 & 74.83 & 67.98 & 93.99 & 69.16 & 66.95 & 93.94 & 82.81 & 83.39 & 75.64 & 78.29  \\
    ReStraV\textsuperscript{\textdagger} \cite{interno2025aigenerated}, NeurIPS 2025                       & 72.57        & 61.39         & \underline{99.09}           & \textbf{99.95}           & \textbf{98.61}           & \textbf{99.57}           & \textbf{100.00}          & 45.01           & \textbf{99.42}           & \underline{98.55}           & 87.42              \\ 
    \rowcolor{gray!5}  
    D3\textsuperscript{\textdagger} \cite{Zheng_2025_ICCV}, ICCV 2025                                      & 90.23        & \underline{93.94}         & 96.25           & 94.88           & 82.22           & 92.38           & 93.21           & 95.60           & 92.50           & 92.52           & 92.37              \\
    \hline \hline
    \rowcolor{gray!25}
    \textbf{ATSS}                                                                                      & \underline{92.22} & \textbf{98.25} & 95.77   & 97.99        & 92.11       & 91.76        & 99.15      & \underline{98.79} & \underline{99.09}     & \textbf{98.62} & \textbf{96.38} \\ 
    \bottomrule
    \end{tabular}
    }
\label{tab:ap_VideoPhy}
\end{table*}

\begin{table*}[!ht]
    \centering
    \caption{Comparison of detection AUC ($\%$) between ATSS and 13 state-of-the-art baselines on VideoPhy.  \textsuperscript{\textdagger} Results are reproduced using the officially released code. \textsuperscript{\textdagger\textdagger} Results are reproduced from our implementation based on the original paper, as no official code is available.}
    \resizebox{\textwidth}{!}{
    \begin{tabular}{c || c c c c c c c c c c | c}
    \toprule
    \rowcolor{gray!25}
    Method                                                                                                 & CVX    & CVX-5B & DM   & Gen-2   & LaVie   & OpenSora      & Pika    & SVD-T2I2V   & VC2   & ZS    & mean   \\
    \midrule
    \hline
    \rowcolor{gray!5}
    STIL\textsuperscript{\textdagger} \cite{10.1145/3474085.3475508}, ACM MM 2021                          & 71.73        & 73.94        & 73.01           & 80.75   & 68.66   & 63.42        & 98.47 & 77.82       & 75.18           & 55.85        & 73.88        \\
    FTCN\textsuperscript{\textdagger} \cite{zheng2021exploring}, ICCV 2021                                 & 80.91        & 81.01        & 70.20           & 95.21   & 72.88   & 76.37        & 95.90 & 87.07       & 86.53           & 74.88        & 82.10        \\
    \rowcolor{gray!5}
    X-CLIP\textsuperscript{\textdagger} \cite{ni2022expanding}, ECCV 2022                                  & 88.99        & 84.84        & 78.57           & 93.35   & 76.51   & \underline{92.68}        & 96.22 & 84.92       & 86.27           & 88.06        & 86.88        \\ 
    TALL\textsuperscript{\textdagger} \cite{xu2023tall}, ICCV 2023                                         & 30.05        & 46.36        & 56.54           & 64.80   & 32.94   & 24.82        & 63.65    & 45.40       & 42.48           & 39.44        & 44.65        \\
    \rowcolor{gray!5}
    FID\textsuperscript{\textdagger} \cite{NEURIPS2024_6dddcff5}, NeurIPS 2024                             & \textbf{93.37} & 91.72        & 97.62   & 98.11 & \underline{96.03} & 86.03 & 99.59 & 95.81 & 95.68     & 89.38 & \underline{94.33} \\
    NPR\textsuperscript{\textdagger} \cite{Tan_2024_CVPR}, CVPR 2024                                       & 72.10        & 73.60        & \textbf{99.70}  & \underline{99.80} & 42.90   & 83.50        & \underline{99.80} & \textbf{99.50}       & 47.20           & 52.90        & 77.10        \\
    \rowcolor{gray!5}
    MINTIME\textsuperscript{\textdagger} \cite{10547206}, TIFS 2024                                       & 85.08        & 87.24        & 82.06           & 86.32          & 76.29   & 89.92        & 93.61          & 80.36                & 85.09           & 89.30        & 85.53        \\ 
    AIGVDet\textsuperscript{\textdagger} \cite{bai2024ai}, PRCV 2024                                       & 78.13        & 67.05        & 68.40           & 75.87   & 71.25   & 69.57        & 97.83    & 72.87       & 81.36           & 77.15        & 75.95        \\
    \rowcolor{gray!5}
    DeMamba\textsuperscript{\textdagger} \cite{chen2024demamba}, arXiv 2024                                & 59.65        & 61.44        & 98.42           & 99.64   & 63.66   & 51.32        & 99.19    & 97.72 & 63.57           & 46.56        & 74.12        \\
    DeCoF\textsuperscript{\textdagger\textdagger} \cite{11210049}, ICME 2025                               & 53.29             & 60.92              & 96.21              & 98.64              & 82.43             & 64.90              & 99.10                 & 92.99               & 84.98                & 63.71                  & 79.72               \\
    \rowcolor{gray!5}
    NSG-VD\textsuperscript{\textdagger} \cite{zhang2025NSGVD}, NeurIPS 2025            & 80.91 & 81.01 & 70.20 & 95.21 & 72.88 & 76.37 & 95.90 & 87.07 & 86.53 & 74.88 & 82.10  \\
    ReStraV\textsuperscript{\textdagger} \cite{interno2025aigenerated}, NeurIPS 2025                       & 78.09             & 65.21              & \underline{98.83}              & \textbf{99.94}              & \textbf{98.91}             & \textbf{99.65}             & \textbf{100.00}             & 41.02             & \textbf{99.44}             & \textbf{99.01}             & 88.01              \\ 
    \rowcolor{gray!5}
    D3\textsuperscript{\textdagger} \cite{Zheng_2025_ICCV}, ICCV 2025                                      & \underline{90.05}           & \underline{93.51}           & 95.56           & 94.34           & 84.26           & 91.90           & 92.02           & 95.09           & 91.68           & 90.91           & 91.93            \\  
    \hline \hline
    \rowcolor{gray!25}
    \textbf{ATSS}                                                                                      & 89.13 & \textbf{97.64} & 94.48           & 97.46   & 89.80 & 90.11 & 99.09 & \underline{98.48}       & \underline{98.87}     & \underline{98.39} & \textbf{95.34} \\ 
    \bottomrule
    \end{tabular}
    }
\label{tab:auc_VideoPhy}
\end{table*}

\begin{table*}[th]
    \centering
    \caption{Comparison of detection AP ($\%$) between ATSS and 13 state-of-the-art baselines on VidProM.  \textsuperscript{\textdagger} Results are reproduced using the officially released code. \textsuperscript{\textdagger\textdagger} Results are reproduced from our implementation based on the original paper, as no official code is available.}
    \resizebox{0.7\textwidth}{!}{
    \begin{tabular}{c || c c c c c c | c}
    \toprule
    \rowcolor{gray!25}
    Method                                                                                                 & MS   & OpenSora   & Pika        & ST2V     &  T2VZ    & VC2     & mean \\
    \midrule
    \hline
    \rowcolor{gray!5}
    STIL\textsuperscript{\textdagger} \cite{10.1145/3474085.3475508}, ACM MM 2021                          & 43.68       & 63.15       & 95.28       & 55.46       & 48.68       & 61.49       & 61.29       \\
    FTCN\textsuperscript{\textdagger} \cite{zheng2021exploring}, ICCV 2021                                 & 74.65       & 90.05       & 94.26       & 83.53       & 46.25       & 87.27       & 79.33       \\
    \rowcolor{gray!5}
    X-CLIP\textsuperscript{\textdagger} \cite{ni2022expanding}, ECCV 2022                                  & 82.24       & 93.47       & 95.80       & 86.42       & 70.92       & 91.60       & 88.05       \\ 
    TALL\textsuperscript{\textdagger} \cite{xu2023tall}, ICCV 2023                                         & 50.93       & 54.50       & 63.47       & 51.50       & 60.99       & 59.70       & 56.85       \\
    \rowcolor{gray!5}
    FID\textsuperscript{\textdagger} \cite{NEURIPS2024_6dddcff5}, NeurIPS 2024                             & 91.35 & 87.68       & 99.59 & \textbf{97.87} & 68.51       & 85.92 & 88.49 \\
    NPR\textsuperscript{\textdagger} \cite{Tan_2024_CVPR}, CVPR 2024                                       & 87.04       & 89.85 & \textbf{99.98} & \underline{89.88} & 88.93 & 70.79       & 87.75       \\
    \rowcolor{gray!5}
    MINTIME\textsuperscript{\textdagger} \cite{10547206}, TIFS 2024                                       & 76.09       & 88.32       & 94.19       & 50.03       & 65.06       & 86.77       & 76.74       \\ 
    AIGVDet\textsuperscript{\textdagger} \cite{bai2024ai}, PRCV 2024                                       & 63.33       & 62.12       & 66.07       & 55.46       & 63.49       & 52.15       & 60.44       \\
    \rowcolor{gray!5}
    DeMamba\textsuperscript{\textdagger} \cite{chen2024demamba}, arXiv 2024                                & \textbf{99.24}       & 48.00       & 86.95       & 33.94       & \textbf{98.47}       & \underline{98.38}       & 77.50       \\
    DeCoF\textsuperscript{\textdagger\textdagger} \cite{11210049}, ICME 2025                               & 87.93       & 85.42       & 99.56       & 76.27       & 93.72       & 96.75       & \underline{90.11}       \\
    \rowcolor{gray!5}
    NSG-VD\textsuperscript{\textdagger} \cite{zhang2025NSGVD}, NeurIPS 2025                                & 74.65 & 90.05 & 94.26 & 83.53 & 46.25 & 87.27 & 79.33  \\
    ReStraV\textsuperscript{\textdagger} \cite{interno2025aigenerated}, NeurIPS 2025                       & \underline{96.99}       & \underline{97.20}       & \underline{99.85}       & 63.96       & 39.63       & 69.14       & 77.80         \\ 
    \rowcolor{gray!5}
    D3\textsuperscript{\textdagger} \cite{Zheng_2025_ICCV}, ICCV 2025                                      & 87.61       & 90.73       & 93.17       & 81.00       & 61.93       & 93.97       & 84.73         \\ 
    \hline \hline 
    \rowcolor{gray!25}
    \textbf{ATSS}                                                                                      & 93.49 & \textbf{98.37} & 99.59 & 81.25       & \underline{97.93} & \textbf{99.50} & \textbf{95.02} \\ 
    \bottomrule
    \end{tabular}
    }
\label{tab:ap_VidProM}
\end{table*}

\begin{table*}[th]
    \centering
    \caption{Comparison of detection AUC ($\%$) between ATSS and 13 state-of-the-art baselines on VidProM. \textsuperscript{\textdagger} Results are reproduced using the officially released code. \textsuperscript{\textdagger\textdagger} Results are reproduced from our implementation based on the original paper, as no official code is available.}
    \resizebox{0.7\textwidth}{!}{
    \begin{tabular}{c || c c c c c c | c}
    \toprule
    \rowcolor{gray!25}
    Method                                                                                                 & MS           & OpenSora    & Pika        & ST2V     &  T2VZ    & VC2     & mean \\
    \midrule
    \hline
    \rowcolor{gray!5}
    STIL\textsuperscript{\textdagger} \cite{10.1145/3474085.3475508}, ACM MM 2021                          & 41.05        & 72.01       & 97.42       & 67.12       & 47.66       & 67.98       & 65.54         \\
    FTCN\textsuperscript{\textdagger} \cite{zheng2021exploring}, ICCV 2021                                 & 70.98        & 90.76       & 95.23       & 87.81       & 50.70       & 88.43       & 80.65         \\
    \rowcolor{gray!5}
    X-CLIP\textsuperscript{\textdagger} \cite{ni2022expanding}, ECCV 2022                                  & 80.59        & 93.02       & 95.85       & 87.47       & 75.86       & 91.24       & 87.34         \\ 
    TALL\textsuperscript{\textdagger} \cite{xu2023tall}, ICCV 2023                                         & 45.29        & 56.39       & 66.91       & 50.31       & 57.80       & 53.48       & 55.03         \\
    \rowcolor{gray!5}
    FID\textsuperscript{\textdagger} \cite{NEURIPS2024_6dddcff5}, NeurIPS 2024                             & 89.86 & 87.36       & 99.62 & \underline{98.10} & 67.01    & 85.65 & 87.93 \\
    NPR\textsuperscript{\textdagger} \cite{Tan_2024_CVPR}, CVPR 2024                                       & 82.61        & \textbf{98.56} & \underline{99.84} & \textbf{98.92} & 93.32 & 56.70    & 88.33         \\ 
    \rowcolor{gray!5}   
    MINTIME\textsuperscript{\textdagger} \cite{10547206}, TIFS 2024                                       & 79.40        & 88.99       & 94.81       & 54.21       & 70.92       & 88.96       & 79.55         \\ 
    AIGVDet\textsuperscript{\textdagger} \cite{bai2024ai}, PRCV 2024                                       & 60.11        & 45.27       & 48.75       & 34.14       & 59.96       & 46.21       & 49.07         \\
    \rowcolor{gray!5}
    DeMamba\textsuperscript{\textdagger} \cite{chen2024demamba}, arXiv 2024                                & 54.44        & 84.02       & 99.29       & 84.94       & 76.34       & 78.61       & 79.61         \\
    DeCoF\textsuperscript{\textdagger\textdagger} \cite{11210049}, ICME 2025                               & 89.18        & 86.98       & 99.56       & 80.14       & \underline{93.85}       & \underline{96.94}       & \underline{90.77}       \\
    \rowcolor{gray!5}
    NSG-VD\textsuperscript{\textdagger} \cite{zhang2025NSGVD}, NeurIPS 2025            & 70.98 & 90.76 & 95.23 & 87.81 & 50.70 & 88.43 & 80.65  \\
    ReStraV\textsuperscript{\textdagger} \cite{interno2025aigenerated}, NeurIPS 2025                       & \textbf{97.96}        & 97.53       & \textbf{99.87}       & 68.44       & 28.60       & 73.79       & 77.70         \\  
    \rowcolor{gray!5} 
    D3\textsuperscript{\textdagger} \cite{Zheng_2025_ICCV}, ICCV 2025                                      & 87.21        & 91.29       & 92.24       & 79.74       & 56.31       & 92.77       & 83.26         \\
    \hline \hline
    \rowcolor{gray!25}
    \textbf{ATSS}                                                                                      & \underline{91.05} & \underline{97.56} & 99.44 & 76.20       & \textbf{97.20} & \textbf{99.25} & \textbf{93.45} \\ 
    \bottomrule
    \end{tabular}
    }
\label{tab:auc_VidProM}
\end{table*}

\begin{table}[ht]
    \centering
    \caption{Comparison of average detection ACC (\%) of ATSS and 13 state-of-the-art baselines on the GenVideo, EvalCrafter, VideoPhy, and VidProM datasets. \textsuperscript{\textdagger} Results are reproduced using the officially released code. \textsuperscript{\textdagger\textdagger} Results are reproduced from our implementation based on the original paper, as no official code is available.}
    \resizebox{\columnwidth}{!}{
    \begin{tabular}{c || c c c c }
    \toprule
    \rowcolor{gray!25}
    Method                                                                                                 & GenVideo        & EvalCrafter     & VideoPhy         & VidProM    \\
    \midrule
    \hline
    \rowcolor{gray!5}
    STIL\textsuperscript{\textdagger} \cite{10.1145/3474085.3475508}, ACM MM 2021                          & 59.90           & 66.78           & 56.62            & 56.58      \\
    FTCN\textsuperscript{\textdagger} \cite{zheng2021exploring}, ICCV 2021                                 & 70.52           & 73.87           & 62.43            & 67.23      \\
    \rowcolor{gray!5}
    X-CLIP\textsuperscript{\textdagger} \cite{ni2022expanding}, ECCV 2022                                  & 75.34           & 76.82           & 69.26            & 73.37      \\   
    TALL\textsuperscript{\textdagger} \cite{xu2023tall}, ICCV 2023                                         & 57.47           & 58.42           & 48.78            & 54.02      \\  
    \rowcolor{gray!5}
    FID\textsuperscript{\textdagger} \cite{NEURIPS2024_6dddcff5}, NeurIPS 2024                             & 54.57           & 63.59           & 65.01            & 54.44      \\  
    NPR\textsuperscript{\textdagger} \cite{Tan_2024_CVPR}, CVPR 2024                                       & 65.41           & 71.36           & 57.00            & 68.04      \\
    \rowcolor{gray!5}
    MINTIME\textsuperscript{\textdagger} \cite{10547206}, TIFS 2024                                       & 78.55           & 81.52           & \underline{77.13}            & 71.74      \\  
    AIGVDet\textsuperscript{\textdagger} \cite{bai2024ai}, PRCV 2024                                       & 49.07           & 57.62           & 53.33            & 47.25      \\  
    \rowcolor{gray!5}
    DeMamba\textsuperscript{\textdagger} \cite{chen2024demamba}, arXiv 2024                                & 54.12           & 62.45           & 42.29            & 42.59      \\ 
    DeCoF\textsuperscript{\textdagger\textdagger} \cite{11210049}, ICME 2025                               & \underline{87.60}           & \underline{89.75}  & 63.71  & \underline{85.47}      \\
    \rowcolor{gray!5}
    NSG-VD\textsuperscript{\textdagger} \cite{zhang2025NSGVD}, NeurIPS 2025                                & 70.52           & 73.87           & 62.43            & 67.23      \\
    ReStraV\textsuperscript{\textdagger} \cite{interno2025aigenerated}, NeurIPS 2025                       & 56.17           & 63.93           & 62.00            & 57.13      \\
    \rowcolor{gray!5}   
    D3\textsuperscript{\textdagger} \cite{Zheng_2025_ICCV}, ICCV 2025                                      & 76.19           & 76.88           & 72.84            & 64.16      \\  
    \hline \hline
    \rowcolor{gray!25}
    \textbf{ATSS}                                                                                      & \textbf{94.32}  & \textbf{96.01}  & \textbf{89.61}   & \textbf{88.42}      \\   
    \bottomrule
    \end{tabular}
    }
\label{tab:acc_all}
\end{table}

\subsection{Evaluation Metrics}
To comprehensively evaluate the performance of the proposed ATSS framework and ensure consistency with previous studies, we employ three widely used quantitative metrics: Average Precision (AP), Area Under the ROC Curve (AUC), and Accuracy (ACC). Among them, AP and AUC serve as the primary evaluation metrics, while ACC is reported as a complementary indicator. This evaluation protocol aligns with recent AI-generated video detection works, including DeCoF~\cite{11210049}, AIGVDet~\cite{bai2024ai}, DeMamba~\cite{chen2024demamba}, DuB3D~\cite{ji2024distinguish}, and D3~\cite{Zheng_2025_ICCV}.  
Given that our datasets maintain a balanced positive-to-negative ratio (1:1), \textit{Accuracy} provides a clear and interpretable measure of overall correctness, while \textit{AP} and \textit{AUC} offer threshold-independent assessments of model discriminability and generalization.  
Following prior works, the decision threshold for accuracy computation is fixed at 0.5. For image-based methods, frame-level results are averaged to produce the corresponding video-level predictions.

\noindent\textbf{Confusion-matrix Definition:}
Let $\mathrm{TP}$, $\mathrm{TN}$, $\mathrm{FP}$, and $\mathrm{FN}$ denote true positives, true negatives, false positives, and false negatives, respectively. Precision (P) and Recall (R) are defined as:
$\mathrm{P} = \mathrm{TP} / (\mathrm{TP}+\mathrm{FP})$,
$\mathrm{R} = \mathrm{TP} / (\mathrm{TP}+\mathrm{FN})$.

\noindent\textbf{Accuracy (ACC):} 
Measures the proportion of correctly classified samples, including both real and AI-generated videos. 
\begin{equation}
\mathrm{ACC}=\frac{\mathrm{TP}+\mathrm{TN}}{\mathrm{TP}+\mathrm{TN}+\mathrm{FP}+\mathrm{FN}}
\end{equation}

\noindent\textbf{Average Precision (AP):} 
Quantifies the area under the precision–recall curve, reflecting the trade-off between precision and recall across all confidence thresholds:
\begin{equation}
\mathrm{AP}=\int_{0}^{1} \! P(R)\, dR ,
\end{equation}
where $P(R)$ denotes precision as a function of recall. In practice, with predictions sorted in descending order of confidence, a common discrete approximation is
\begin{equation}
\mathrm{AP}=\sum_{k=1}^{K} \big(R_k-R_{k-1}\big)\, P_k ,
\end{equation}
where $\{(P_k,R_k)\}_{k=1}^{K}$ denotes sampled points on the PR curve and $R_0=0$.

\noindent\textbf{Area Under the ROC Curve (AUC):} 
Evaluates the separability between real and AI-generated videos by integrating the trade-off between true positive and false positive rates. 
Let $\mathrm{TPR}(\theta) = \mathrm{TP}(\theta) / (\mathrm{TP}(\theta)+\mathrm{FN}(\theta))$ and $\mathrm{FPR}(\theta) = \mathrm{FP}(\theta) / (\mathrm{FP}(\theta)+\mathrm{TN}(\theta))$ be the true- and false-positive rates at threshold $\theta$. The AUC is defined as follows:
\begin{equation}
\mathrm{AUC}=\int_{0}^{1} \! \mathrm{TPR}(\mathrm{FPR}) \, d\mathrm{FPR},
\end{equation}
which represents the area under the ROC curve parameterized by $(\mathrm{FPR}(\theta),\,\mathrm{TPR}(\theta))$.

\begin{table*}[htbp]
    \centering
    \caption{Ablation performance of individual components in ATSS on GenVideo, reported in terms of AP (\%).}
    \resizebox{\textwidth}{!}{
    \begin{tabular}{c c c c || c c c c c c c c c c | c}
    \toprule
    \rowcolor{gray!25}
    Visual          & Textual         & Cross-Modal     & Cross-Attentive      & MS            & MPS            & MV          & HotShot             & Show-1            & Gen2                & Crafter            & LaVie              & Sora               & WS         & mean   \\
    \midrule
    \hline
    $\checkmark$    & $\times$        & $\times$        & $\times$           & 45.39                 & 44.62                  & 43.70              & 45.61              & 43.76             & 43.12               & 44.18              & 44.24              & 47.47              & 43.84              & 44.59   \\ 
    $\times$        & $\checkmark$    & $\times$        & $\times$           & 85.69                 & 90.30                  & 94.54              & 94.45              & 94.99             & 96.02               & 95.50              & 93.81              & 72.59              & 88.52              & 90.64   \\ 
    $\times$        & $\times$        & $\checkmark$    & $\times$           & 87.17                 & 89.89                  & 95.34              & 95.01              & 93.12             & 95.85               & 95.98              & 93.49              & \textbf{86.05}  & 88.58              & 92.05   \\
    \hline
    $\times$        & $\checkmark$    & $\checkmark$    & $\checkmark$       & 83.85                 & 89.67                  & 88.67              & 94.21              & 93.64             & 93.15               & 94.38              & 89.18              & 82.08              & 88.23              & 89.71   \\ 
    $\checkmark$    & $\times$        & $\checkmark$    & $\checkmark$       & 92.55                 & 97.27                  & 99.08              & 98.82              & 99.36             & \underline{99.38}               & 99.55              & 96.02              & \underline{86.03}              & 97.49              & 96.55   \\ 
    $\checkmark$    & $\checkmark$    & $\times$        & $\checkmark$       & 90.07                 & 95.40                  & 97.29              & 96.04              & 97.22             & 97.09               & 98.42              & 95.26              & 81.77              & 93.11              & 94.17   \\
    \hline
    $\checkmark$    & $\checkmark$    & $\checkmark$    & $\times$           & \underline{93.57}     & \underline{98.47}      & \underline{99.42}  & \underline{99.26}  & \textbf{99.73}    & \textbf{99.68}      & \underline{99.85}  & \underline{97.97}  & 84.76              & \underline{98.28}  & \underline{97.10} \\ 
    \midrule
    \hline
    \rowcolor{gray!25}
    $\checkmark$    & $\checkmark$    & $\checkmark$    & $\checkmark$       & \textbf{95.21}        & \textbf{99.03}         & \textbf{99.68}     & \textbf{99.61}     & \underline{99.72} & \textbf{99.68}      & \textbf{99.94}     & \textbf{98.51}     & 85.75              & \textbf{98.48}     & \textbf{97.56}   \\ 
    \bottomrule
    \end{tabular}
    }
\label{tab:ablation_ap}
\end{table*}

\begin{table*}[htbp]
    \centering
    \caption{Ablation performance of individual components in ATSS on GenVideo, reported in terms of AUC (\%).}
    \resizebox{\textwidth}{!}{
    \begin{tabular}{c c c c || c c c c c c c c c c | c}
    \toprule
    \rowcolor{gray!25}
    Visual          & Textual         & Cross-Modal     & Cross-Attentive      & MS     & MPS     & MV      & HotShot     & Show-1     & Gen2     & Crafter     & LaVie     & Sora     & WS     & mean   \\
    \midrule
    \hline
    $\checkmark$    & $\times$        & $\times$        & $\times$           & 50.75          & 49.90           & 47.57           & 52.03       & 47.87      & 45.75    & 48.48       & 48.88     & 53.44    & 48.10          & 49.28   \\ 
    $\times$        & $\checkmark$    & $\times$        & $\times$           & 84.56          & 89.66           & 93.81           & 93.74       & 94.45      & 95.73    & 94.90       & 93.04     & 72.40    & 87.06          & 89.93   \\ 
    $\times$        & $\times$        & $\checkmark$    & $\times$           & 87.14          & 91.24           & 95.32           & 95.09       & 94.45      & 95.68    & 95.97       & 93.51     & 84.08    & 88.11          & 92.06   \\
    \hline
    $\times$        & $\checkmark$    & $\checkmark$    & $\checkmark$       & 85.22          & 91.53           & 93.47           & 95.73       & 96.26      & 96.03    & 96.42       & 92.39     & 84.24    & 89.14          & 92.04   \\ 
    $\checkmark$    & $\times$        & $\checkmark$    & $\checkmark$       & 90.06          & 96.84           & 98.89           & 98.47       & 99.26      & 99.25    & 99.44       & 95.02     & \textbf{85.85}    & 96.54          & 95.96   \\ 
    $\checkmark$    & $\checkmark$    & $\times$        & $\checkmark$       & 90.04          & 95.39           & 96.93           & 96.73       & 96.93      & 97.37    & 98.21       & 95.34     & 84.28    & 93.21          & 94.44   \\  
    \hline
    $\checkmark$    & $\checkmark$    & $\checkmark$    & $\times$           & \underline{91.99}          & \underline{98.13}           & \underline{99.28}           & \underline{99.06}       & \textbf{99.70}      & \textbf{99.65}    & \underline{99.84}       & \underline{97.39}     & 81.43    & \underline{97.72}          & \underline{96.42}   \\ 
    \hline \hline
    \rowcolor{gray!25}
    $\checkmark$    & $\checkmark$    & $\checkmark$    & $\checkmark$       & \textbf{93.49}          & \textbf{98.82}           & \textbf{99.62}           & \textbf{99.44}        & \underline{99.67}      & \underline{99.58}    & \textbf{99.92}      & \textbf{97.95}     & \underline{84.89}    & \textbf{97.76}   & \textbf{97.11}  \\ 
    \bottomrule
    \end{tabular}
    }
\label{tab:ablation_auc}
\end{table*}

\begin{table*}[htbp]
    \centering
    \caption{Ablation performance of individual components in ATSS on GenVideo, reported in terms of ACC (\%).}
    \resizebox{\textwidth}{!}{
    \begin{tabular}{c c c c || c c c c c c c c c c | c}
    \toprule
    \rowcolor{gray!25}
    Visual          & Textual         & Cross-Modal     & Cross-Attentive      & MS     & MPS     & MV      & HotShot     & Show-1     & Gen2     & Crafter     & LaVie     & Sora     & WS     & mean   \\
    \midrule
    \hline
    $\checkmark$    & $\times$        & $\times$        & $\times$           & 63.02          & 66.27           & 68.26           & 69.84       & 66.03      & 65.06    & 69.38       & 66.07     & 62.75    & 64.75          & 66.14   \\ 
    $\times$        & $\checkmark$    & $\times$        & $\times$           & 76.51          & 81.90           & 85.90           & 86.75       & 86.51      & 88.41    & 87.49       & 84.92     & 66.67    & 77.76          & 82.28   \\ 
    $\times$        & $\times$        & $\checkmark$    & $\times$           & 78.97          & 84.05           & 87.41           & 87.38       & 87.38      & 88.12    & 87.81       & 86.11     & 71.57    & 80.11          & 83.89   \\
    \hline
    $\times$        & $\checkmark$    & $\checkmark$    & $\checkmark$       & 78.41          & 84.29           & 86.88           & 89.21       & 90.00      & 89.90    & 89.04       & 86.07     & \textbf{79.41}    & 82.16          & 85.54   \\ 
    $\checkmark$    & $\times$        & $\checkmark$    & $\checkmark$       & 82.70          & 88.33           & 92.46           & 91.59       & \underline{92.14}      & 93.48    & 92.02       & 87.22     & \underline{78.43}    & 89.47          & 88.78   \\ 
    $\checkmark$    & $\checkmark$    & $\times$        & $\checkmark$       & \underline{83.10}          & 89.13           & 87.68           & 89.21       & 88.25      & 89.33    & 90.23       & 87.70     & 76.47    & 85.44          & 86.65   \\  
    \hline
    $\checkmark$    & $\checkmark$    & $\checkmark$    & $\times$           & 80.24          & \underline{92.46}           & \underline{97.16}           & \underline{96.67}       & \textbf{97.86}      & \underline{97.83}    & \underline{99.09}       & \underline{91.94}     & 60.78    & \underline{92.44}          & \underline{90.65}   \\ 
    \hline \hline
    \rowcolor{gray!25}
    $\checkmark$    & $\checkmark$    & $\checkmark$    & $\checkmark$       & \textbf{87.06}          & \textbf{95.48}           & \textbf{97.78}           & \textbf{97.54}       & \textbf{97.86}      & \textbf{98.91}    & \textbf{99.44}       & \textbf{96.43}     & 76.47    & \textbf{96.22}          & \textbf{94.32}   \\ 
    \bottomrule
    \end{tabular}
    }
\label{tab:ablation_acc}
\end{table*}

\subsection{Quantitative Results}
Tables~\ref{tab:ap_GenVideo} – \ref{tab:acc_all} demonstrate a comprehensive comparison between ATSS and 13 state-of-the-art approaches on the GenVideo, EvalCrafter, VideoPhy, and VidProM benchmarks in terms of AP, AUC, and ACC. 
In these tables, bold and underlined values represent the best and second-best performance, respectively.
To ensure a fair comparison, baseline results are either reproduced using their officially released code, or re-implemented based on the original publications when official code is unavailable.

\subsubsection{GenVideo}
On the GenVideo dataset, which encompasses 10 distinct video generators, ATSS demonstrates superior detection performance across all metrics.
As reported in Table ~\ref{tab:ap_GenVideo}, ATSS achieves a mean AP of 97.56\%, which markedly surpasses the strongest baseline D3 by an absolute margin of 5.40\%. 
Regarding the AUC metric, our approach obtains a mean AUC of 97.11\%, outperforming the best competing approach ReStraV by 4.62\%, as shown in Table ~\ref{tab:auc_GenVideo}. 
In terms of ACC, ATSS improves over the second-best method DeCoF by 6.72\%, as summarized in Table ~\ref{tab:acc_all}. 

However, a performance degradation is observed on the Sora subset, yielding 85.75\% AP and 84.89\% AUC.
This decline arises from Sora’s near-real-world physical dynamics that significantly diminish the ``unnatural temporal self-similarity'' and ``forced alignment'' artifacts that ATSS targets. 
By mimicking the stochastic and heterogeneous dynamics of natural videos, Sora renders the deterministic fingerprints of anchor-driven evolution increasingly indistinguishable from real videos. Such high-fidelity generation poses a substantial challenge to the proposed similarity-based detection framework.

\subsubsection{EvalCrafter}
On the EvalCrafter benchmark, one can see that ATSS outperforms the state-of-the-art method FID by improvements of 3.02\% and 2.93\% in terms of AP and AUC, respectively, as detailed in Tables ~\ref{tab:ap_EvalCrafter} and ~\ref{tab:auc_EvalCrafter}.
Regarding the ACC metric, our approach achieves 6.26\% gains over the second-best baseline DeCoF, as reported in Table ~\ref{tab:acc_all}.

Among these baselines, deepfake video detection methods (e.g., TALL and STIL) exhibit limited efficacy in detecting general generative artifacts.
Interestingly, the image-level detector FID demonstrates comparatively robust generalization. This stems from its focus on local feature artifacts, which enables the model to remain largely invariant to diverse semantic scenes. 
Although AI-generated video detection frameworks like DeCoF provide competitive results by modeling inter-frame temporal consistency, they struggle to capture the complex, multi-faceted similarity patterns inherent in anchor-driven synthesis. This reveals that the triple-similarity representational space of ATSS is essential for capturing the structured cross-modal fingerprints that distinguish AI-generated videos from authentic ones.

\subsubsection{VideoPhy}
On the VideoPhy dataset, ATSS consistently outperforms all state-of-the-art baselines across key metrics. 
As detailed in Tables~\ref{tab:ap_VideoPhy} and~\ref{tab:auc_VideoPhy}, ATSS achieves an average AP of 96.38\% and an average AUC of 95.34\%, surpassing the second-best baseline FID by absolute margins of 1.69\% and 1.01\% in AP and AUC, respectively. 
In terms of detection accuracy (ACC), as reported in Table~\ref{tab:acc_all}, ATSS attains 89.61\% on VideoPhy, delivering a substantial 12.48 percentage points improvement over the second-best baseline MINTIME.

Notably, deepfake video detection methods (e.g., STIL and TALL) struggle to generalize to the physics-rich, non-facial scenarios of VideoPhy, as they are primarily optimized for capturing localized manipulation-induced discrepancies.
Interestingly, the image-level detector FID delivers comparatively strong performance by employing patch-level analysis to mitigate semantic dependencies.
Although D3 achieves competitive results by leveraging second-order temporal statistics, it remains a purely visual statistical approach that fails to fully exploit the structured cross-modal correlations inherent in video synthesis.

\subsubsection{VidProM}
As a million-scale real prompt-gallery benchmark dedicated to text-to-video diffusion models, VidProM features diverse prompts from real users and videos generated by multiple state-of-the-art models. 
On this dataset, ATSS achieves AP and AUC improvements of 4.91\% and 2.68\% over the strongest baseline DeCoF, as illustrated in Table~\ref{tab:ap_VidProM} and Table~\ref{tab:auc_VidProM}, respectively.
Regarding the ACC metric, ATSS improves by 2.95\% over the state-of-the-art method DeCoF, as presented in Table~\ref{tab:acc_all}.
These results demonstrate that, unlike DeCoF, which relies solely on visual features, the adaptive fusion of multimodal features across frames in ATSS is critical for capturing the nuanced artifacts introduced by high‑quality generative video models.

Nevertheless, a notable performance drop is observed on the StreamingT2V (ST2V) subset, where AP and AUC drop to 81.25\% and 76.20\%, respectively.
This performance bottleneck is mainly attributed to the hierarchical refinement strategy of ST2V.
Unlike other video generators that often preserve structured generative noise, ST2V employs an autoregressive refinement stage to explicitly suppress temporal flickering and repetitive self-similarity~\cite{wang2024vidprom}. 
While ATSS targets the deterministic and mechanical smoothness derived from forced anchor alignment, ST2V’s refinement process produces stochastic fluid transitions that more closely resemble the heterogeneous dynamics of real-world videos. 
Consequently, the structured generative fingerprints typically found in similarity matrices are significantly diminished, making the refined motion trajectories of ST2V particularly difficult to differentiate from authentic videos.

\subsection{Ablation Study}
To validate the structural effectiveness and robustness of ATSS, we conduct comprehensive ablation studies on the GenVideo dataset.
First, we analyze the contribution of each core component in ATSS, specifically focusing on the proposed triple-similarity based multimodal representation and the cross-attentive fusion mechanism. 
We compare the performance of ATSS across diverse configurations, including single-branch variants, various multi-branch combinations, and variants ablating the bidirectional cross-attentive fusion module, as detailed in Tables~\ref{tab:ablation_ap}–\ref{tab:ablation_acc}.
Second, we investigate the framework’s robustness with respect to different backbones by evaluating the impact of various image captioning models and visual-textual feature encoders on its overall performance. 
These results, reported in Tables~\ref{tab:ablation2_ap}–\ref{tab:ablation2_acc}, demonstrate that ATSS consistently maintains superior detection performance irrespective of the choice of auxiliary modules.

\subsubsection{Effectiveness of Single-Branch Variants}
We remove two of the three branches each time, evaluating our approach with only the visual branch, only the textual branch, and only the cross‑modal branch active.
The absence of both textual and cross-modal branches leads to the most severe performance degradation, with AP, AUC, and ACC dropping by 52.97\%, 47.83\%, and 28.18\%, respectively.
Ablating the visual and cross-modal branches reduces AP, AUC, and ACC by 6.92\%, 7.18\%, and 12.04\%, respectively.
The removal of visual and textual branches results in drops of 5.51\%, 5.05\%, and 10.43\% in AP, AUC, and ACC.
These results underscore the importance of high-level semantics embedded in textual and cross-modal representations.

\subsubsection{Effectiveness of Multi-Branch Combinations}
We individually ablate one of the three branches to validate the effectiveness of multi-branch combinations.
Removing the visual branch causes noticeable performance declines of 7.85\% in AP, 5.07\% in AUC, and 8.78\% in ACC.
Ablating the textual branch reduces AP, AUC, and ACC by 1.01\%, 1.15\%, and 5.54\%, respectively.
The removal of the cross-modal branch results in drops of 3.39\%, 2.67\%, and 7.67\% in AP, AUC, and ACC.
These experiments suggest that visual features are a pivotal complementary component to the high-level semantics from textual and cross-modal representations.

\subsubsection{Impact of Cross-Attentive Fusion Module}
We further investigate the impact of the cross-attentive fusion mechanism by removing it from our three-branch framework.
It is observed that ablating the cross-attentive fusion module leads to a performance drop of 0.46\% in AP, 0.69\% in AUC, and 3.67\% in ACC.
This experiment reveals that the bidirectional cross-attentive fusion module effectively facilitates feature interactions across different modalities, thereby further improving the detection efficacy of AI-generated video detection.

\begin{table*}[htbp]
    \centering
    \caption{Ablation results of ATSS using various image captioning backbones and visual-textual encoders on the GenVideo dataset, measured by AP (\%).}
    \resizebox{0.95\textwidth}{!}{
    \begin{tabular}{c c || c c c c c c c c c c | c}
    \toprule
    \rowcolor{gray!25}
    Caption Model               & Visual \& Textual Encoder    & MS       & MPS       & MV       & HotShot       & Show-1       & Gen2       & Crafter       & Lavie       & Sora       & WS       & mean    \\
    \midrule
    \hline
    \multirow{5}{*}{BLIP-base}  & CLIP-P16                     & 92.25    & 97.25     & 98.69    & 98.56         & 99.32        & 99.18      & 99.56         & 95.49       & 84.90      & 96.00    & 96.12   \\ 
                                & CLIP-P32                     & 94.00    & 97.70     & 99.02    & 98.69         & 99.25        & 99.22      & 99.63         & 97.31       & \textbf{87.80}      & 97.07    & 96.97   \\ 
                                & XCLIP-P16                    & 76.37    & 79.21     & 80.76    & 82.16         & 82.66        & 83.29      & 83.77         & 79.08       & 63.37      & 83.25    & 79.39   \\ 
                                & XCLIP-P32                    & 91.92    & 96.68     & 99.09    & 98.23         & 98.72        & 99.35      & 99.15         & 97.94       & 72.66      & 95.57    & 94.93   \\ 
    \hline
    \multirow{5}{*}{BLIP-large} & CLIP-P16                     & 86.86    & 93.35     & 94.76    & 94.09         & 95.89        & 94.27      & 96.16         & 94.15       & 73.97      & 88.62    & 91.21   \\ 
                                & CLIP-P32                     & 82.97    & 89.53     & 93.64    & 92.36         & 91.87        & 93.07      & 93.72         & 90.50       & 74.32      & 86.47    & 88.84   \\ 
                                & XCLIP-P16                    & 90.65    & 97.10     & 99.37    & 96.71         & 98.43        & 98.97      & 99.18         & 97.20       & 82.57      & 94.43    & 95.46   \\ 
                                & XCLIP-P32                    & 94.53    & 97.45     & 99.43    & 99.00         & 98.69        & \underline{99.70}      & 99.68         & 98.36       & 85.04      & 96.10    & 96.80   \\ 
    \hline
    \multirow{5}{*}{BLIP2-t5}   & CLIP-P16                     & \textbf{95.76}    & \textbf{99.27}     & 99.35    & 99.55         & \underline{99.70}        & 99.61      & \underline{99.84}         & 98.56       & 78.62      & \underline{98.00}    & 96.82   \\ 
                                & CLIP-P32                     & 93.51    & 97.93     & 98.83    & 98.96         & 98.98        & 99.08      & 99.58         & 97.47       & 80.17      & 96.18    & 96.07   \\ 
                                & XCLIP-P16                    & 93.89    & 97.71     & 99.27    & 98.97         & 99.05        & 99.50      & 99.71         & \underline{98.99}       & 75.92      & 96.99    & 96.00   \\ 
                                & XCLIP-P32                    & 94.94    & 98.46     & \underline{99.52}    & \underline{99.60}         & 99.52        & \textbf{99.94}      & 99.83         & \textbf{99.25}       & 84.13      & 97.47    & \underline{97.27}   \\ 
                                & BLIP2-ITM                    & \underline{95.21}     & \underline{99.03}     & \textbf{99.68}      & \textbf{99.61}     & \textbf{99.72}    & 99.68     & \textbf{99.94}        & 98.51      & \underline{85.75}    & \textbf{98.48}         & \textbf{97.56}      \\ 
    \bottomrule
    \end{tabular}
    }
\label{tab:ablation2_ap}
\end{table*}

\begin{table*}[htbp]
    \centering
    \caption{Ablation results of ATSS using various image captioning backbones and visual-textual encoders on the GenVideo dataset, measured by AUC (\%).}
    \resizebox{0.95\textwidth}{!}{
    \begin{tabular}{c c || c c c c c c c c c c | c}
    \toprule
    \rowcolor{gray!25}
    Caption Model               & Visual \& Textual Encoder    & MS       & MPS       & MV       & HotShot       & Show-1       & Gen2       & Crafter       & Lavie       & Sora       & WS       & mean    \\
    \midrule
    \hline
    \multirow{5}{*}{BLIP-base}  & CLIP-P16                     & 90.09    & 96.86     & 98.22    & 98.25         & 99.21        & 98.94      & 99.45         & 93.53       & 83.20      & 94.65    & 95.24   \\ 
                                & CLIP-P32                     & 93.15    & 97.74     & 98.91    & 98.69         & 99.33        & 99.15      & 99.62         & 96.50       & \textbf{87.77}      & 96.24    & 96.71   \\ 
                                & XCLIP-P16                    & 83.08    & 86.75     & 89.55    & 89.61         & 90.10        & 90.14      & 91.73         & 86.04       & 71.32      & 88.43    & 86.67   \\ 
                                & XCLIP-P32                    & 90.30    & 96.04     & 98.92    & 98.84         & 98.64        & 99.43      & 99.02         & 97.32       & 76.66      & 94.40    & 94.96   \\ 
    \hline
    \multirow{5}{*}{BLIP-large} & CLIP-P16                     & 85.25    & 91.50     & 93.74    & 93.42         & 95.28        & 93.02      & 95.20         & 93.33       & 69.24      & 86.43    & 89.64   \\ 
                                & CLIP-P32                     & 85.06    & 91.01     & 95.46    & 93.72         & 93.86        & 94.54      & 95.47         & 91.14       & 78.66      & 88.51    & 90.74   \\ 
                                & XCLIP-P16                    & 90.96    & 97.06     & 99.34    & 97.70         & 98.75        & 98.96      & 99.14         & 96.80       & 80.78      & 94.17    & 95.37   \\ 
                                & XCLIP-P32                    & 93.54    & 97.28     & 99.31    & 99.00         & 98.74        & \underline{99.69}      & 99.64         & 97.95       & 82.89      & 95.10    & 96.31   \\ 
    \hline
    \multirow{5}{*}{BLIP2-t5}   & CLIP-P16                     & \textbf{94.19}    & \textbf{99.11}     & 99.17    & 99.40         & \underline{99.61}        & 99.44      & 99.79         & 97.88       & 82.58      & \underline{97.09}    & 96.83   \\ 
                                & CLIP-P32                     & 91.70    & 97.49     & 98.44    & 98.60         & 98.81        & 98.79      & 99.46         & 96.59       & 82.78      & 94.74    & 95.74   \\ 
                                & XCLIP-P16                    & 93.13    & 97.86     & 99.24    & 98.81         & 99.08        & 99.51      & 99.70         & \underline{98.85}       & 73.39      & 96.83    & 95.64   \\ 
                                & XCLIP-P32                    & \underline{94.04}    & 98.18     & \underline{99.38}    & \textbf{99.55}         & 99.46        & \textbf{99.94}      & \underline{99.83}         & \textbf{99.22}       & 83.31      & 96.98    & \underline{96.99}   \\ 
                                & BLIP2-ITM                    & 93.49 & \underline{98.82}  & \textbf{99.62}      & \underline{99.44}    & \textbf{99.67}    & 99.58       & \textbf{99.92}          & 97.95        & \underline{84.89}                & \textbf{97.76}             & \textbf{97.11}  \\ 
    \bottomrule
    \end{tabular}
    }
\label{tab:ablation2_auc}
\end{table*}

\begin{table*}[htbp]
    \centering
    \caption{Ablation results of ATSS using various image captioning backbones and visual-textual encoders on the GenVideo dataset, measured by ACC (\%).}
    \resizebox{0.95\textwidth}{!}{
    \begin{tabular}{c c || c c c c c c c c c c | c}
    \toprule
    \rowcolor{gray!25}
    Caption Model               & Visual \& Textual Encoder    & MS       & MPS       & MV       & HotShot       & Show-1       & Gen2       & Crafter       & Lavie       & Sora       & WS       & mean    \\
    \midrule
    \hline
    \multirow{5}{*}{BLIP-base}  & CLIP-P16                     & 82.86    & 87.70     & 89.45    & 89.52         & 90.71        & 90.18      & 90.43         & 86.07       & \textbf{77.45}      & 86.62    & 87.10   \\ 
                                & CLIP-P32                     & 84.84    & 92.86     & 96.19    & 95.24         & 96.83        & 96.78      & 97.26         & 91.67       & 73.53      & 91.26    & 91.64   \\ 
                                & XCLIP-P16                    & 74.52    & 76.03     & 76.33    & 75.63         & 75.16        & 76.41      & 77.24         & 74.72       & 64.71      & 76.58    & 74.73   \\ 
                                & XCLIP-P32                    & 80.63    & 89.68     & 94.68    & 95.95         & 93.41        & 96.70      & 96.07         & 92.14       & 64.71      & 85.44    & 88.94   \\ 
    \hline
    \multirow{5}{*}{BLIP-large} & CLIP-P16                     & 76.27    & 80.40     & 81.83    & 81.75         & 84.13        & 81.44      & 83.76         & 83.10       & 68.63      & 77.63    & 79.89   \\ 
                                & CLIP-P32                     & 76.03    & 81.59     & 89.18    & 86.19         & 86.35        & 87.44      & 88.24         & 83.45       & 71.57      & 79.12    & 82.92   \\ 
                                & XCLIP-P16                    & 83.17    & 89.76     & 94.15    & 93.33         & 93.73        & 93.44      & 93.09         & 90.32       & \textbf{77.45}      & 86.74    & 89.52   \\ 
                                & XCLIP-P32                    & 82.46    & 91.35     & 96.45    & 96.03         & 93.97        & 97.67      & 97.30         & 92.46       & 69.61      & 86.00    & 90.33   \\ 
    \hline
    \multirow{5}{*}{BLIP2-t5}   & CLIP-P16                     & \textbf{87.14}    & \textbf{95.95}     & 94.86    & \textbf{97.62}         & \underline{97.46}        & 98.43      & \underline{98.69}         & 94.92       & 64.71      & \underline{92.44}    & \underline{92.22}   \\ 
                                & CLIP-P32                     & 83.49    & 87.62     & 89.45    & 89.29         & 88.49        & 90.70      & 91.30         & 87.90       & \underline{76.47}      & 86.43    & 87.11   \\ 
                                & XCLIP-P16                    & 84.92    & 92.06     & 95.21    & 94.29         & 95.63        & 96.58      & 96.58         & 94.72       & 62.75      & 90.09    & 90.28   \\ 
                                & XCLIP-P32                    & 81.43    & 93.33     & \underline{96.54}    & 97.46         & 96.75        & \textbf{99.07}      & 98.65         & \underline{96.15}       & 59.80      & 88.60    & 90.78   \\ 
                                & BLIP2-ITM                    & \underline{87.06}          & \underline{95.48}           & \textbf{97.78}           & \underline{97.54}       & \textbf{97.86}      & \underline{98.91}    & \textbf{99.44}       & \textbf{96.43}     & \underline{76.47}    & \textbf{96.22}          & \textbf{94.32}   \\
    \bottomrule
    \end{tabular}
    }
\label{tab:ablation2_acc}
\end{table*}

\begin{figure*}[!ht]
    \centering
    \subfloat[Crafter]{\includegraphics[width=0.19\textwidth]{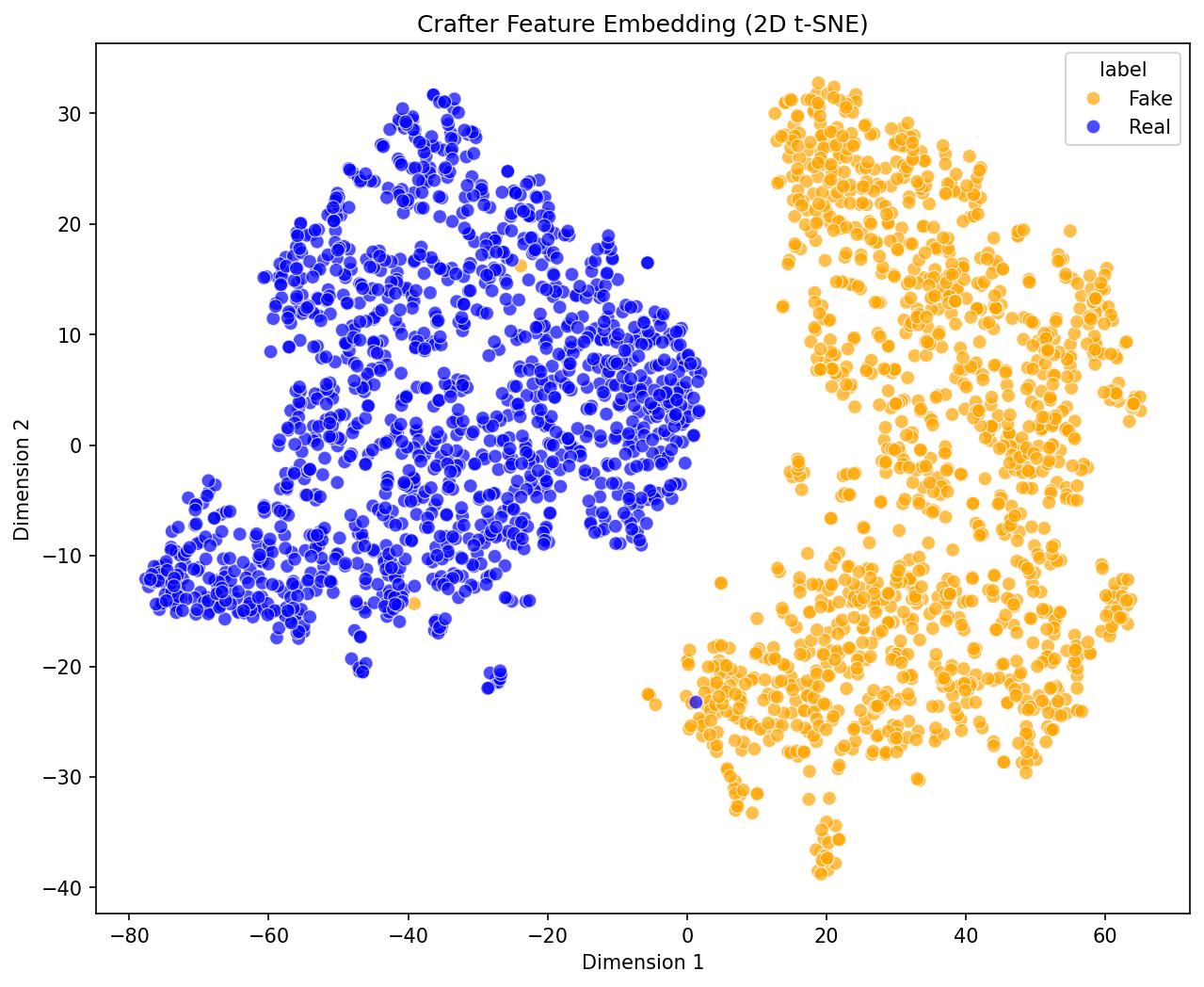}}\hfill
    \subfloat[Gen2]{\includegraphics[width=0.19\textwidth]{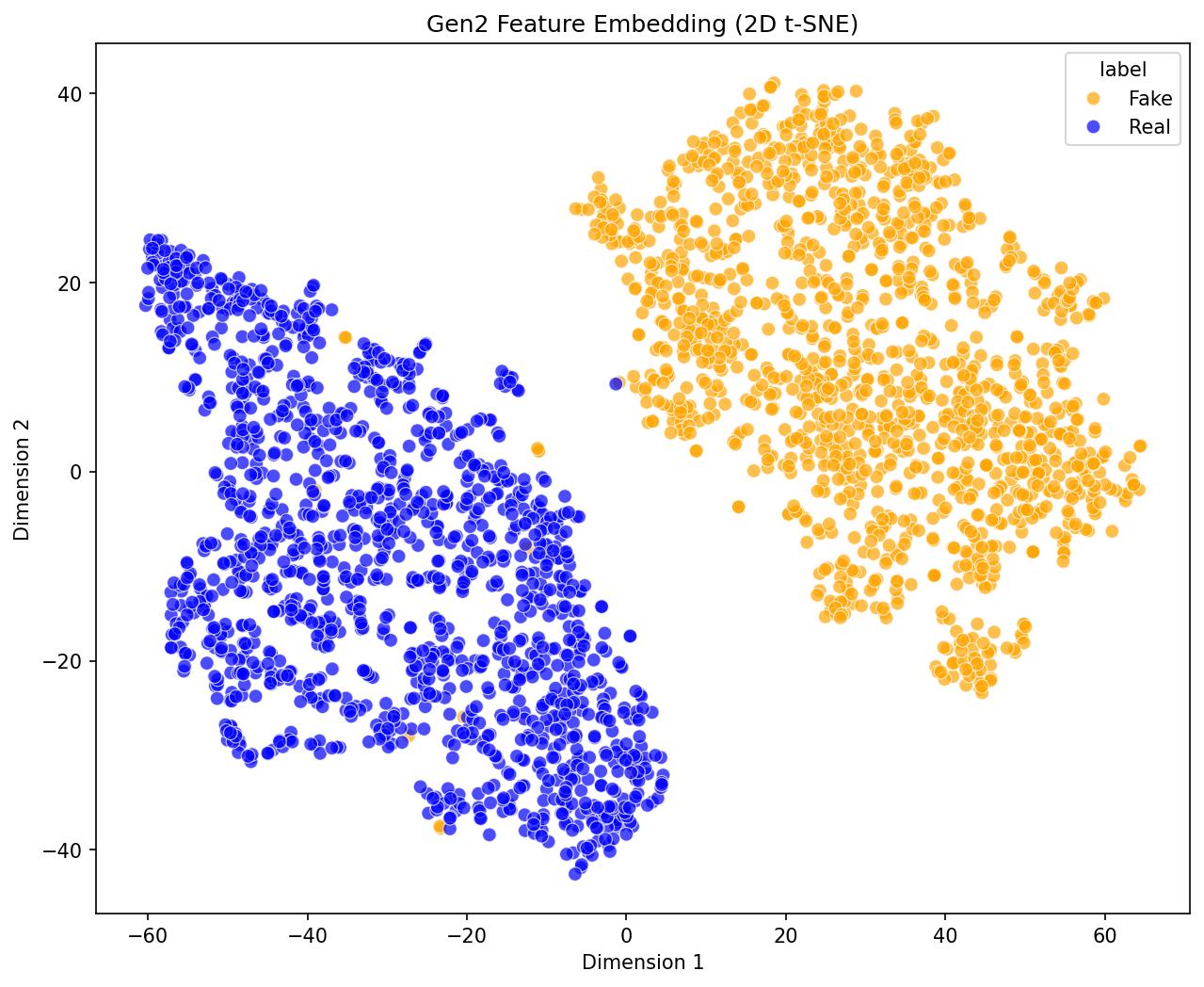}}\hfill
    \subfloat[HotShot]{\includegraphics[width=0.19\textwidth]{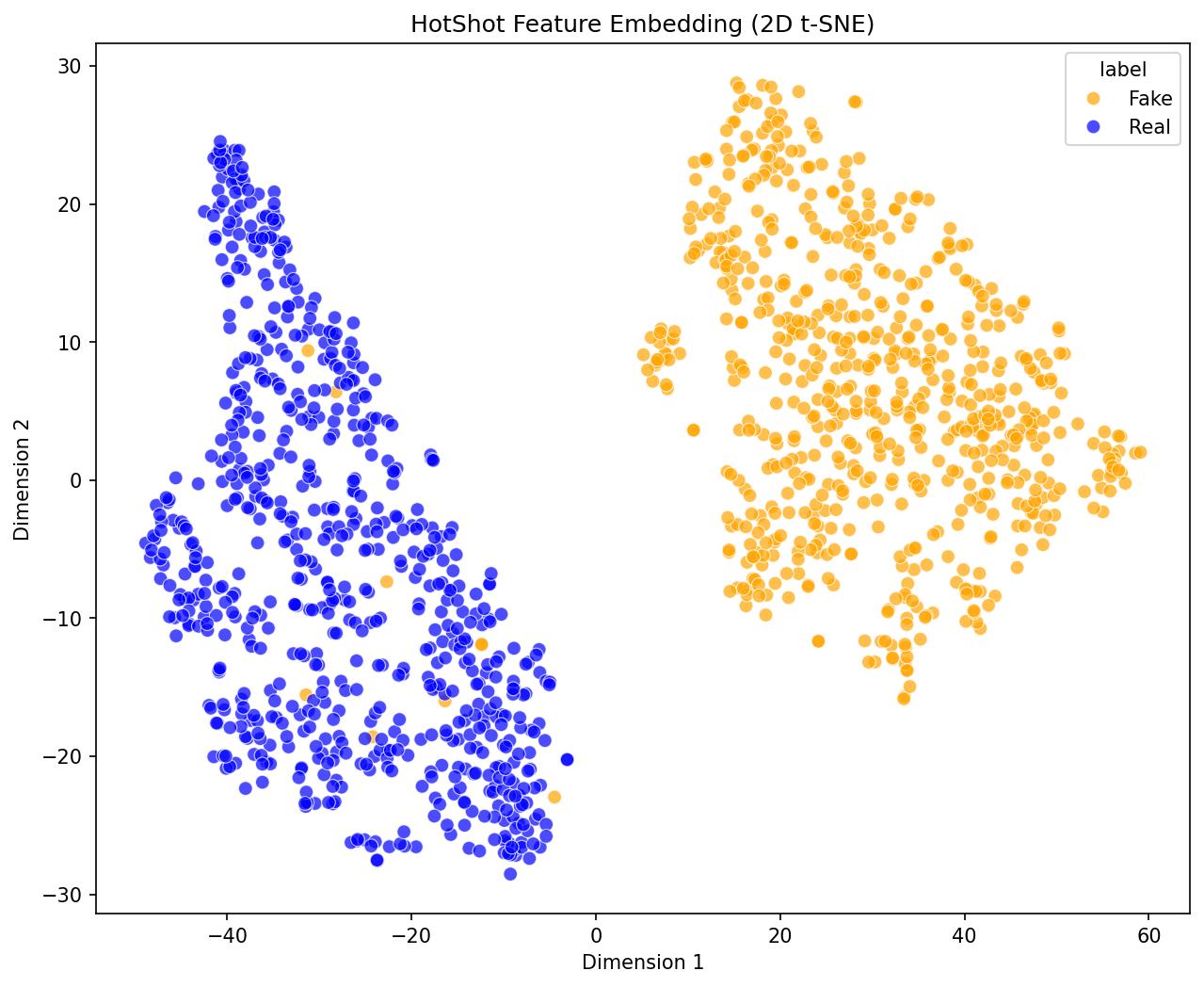}}\hfill
    \subfloat[Lavie]{\includegraphics[width=0.19\textwidth]{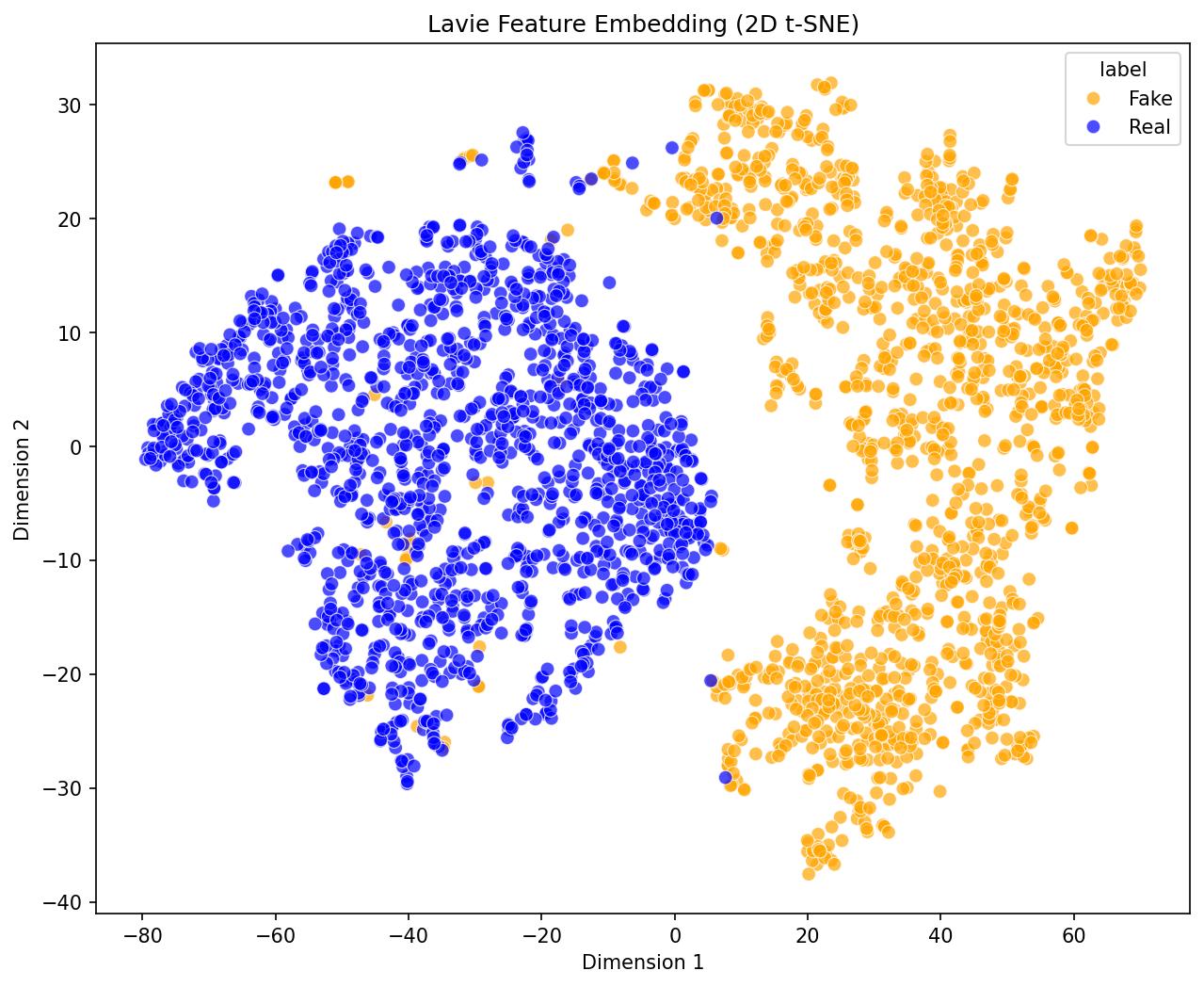}}\hfill
    \subfloat[ModelScope]{\includegraphics[width=0.19\textwidth]{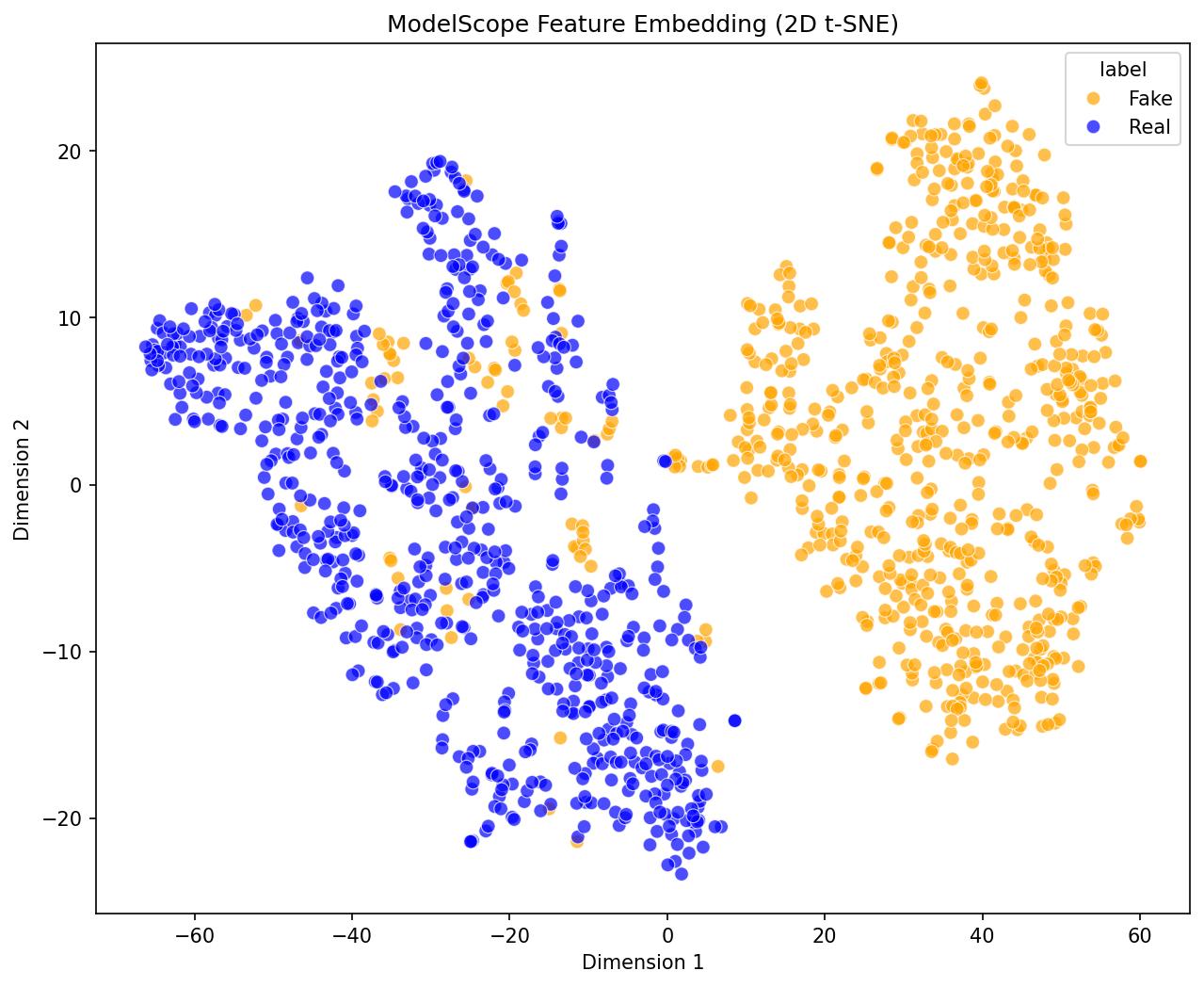}}\\[-0.2em]
    \subfloat[MoonValley]{\includegraphics[width=0.19\textwidth]{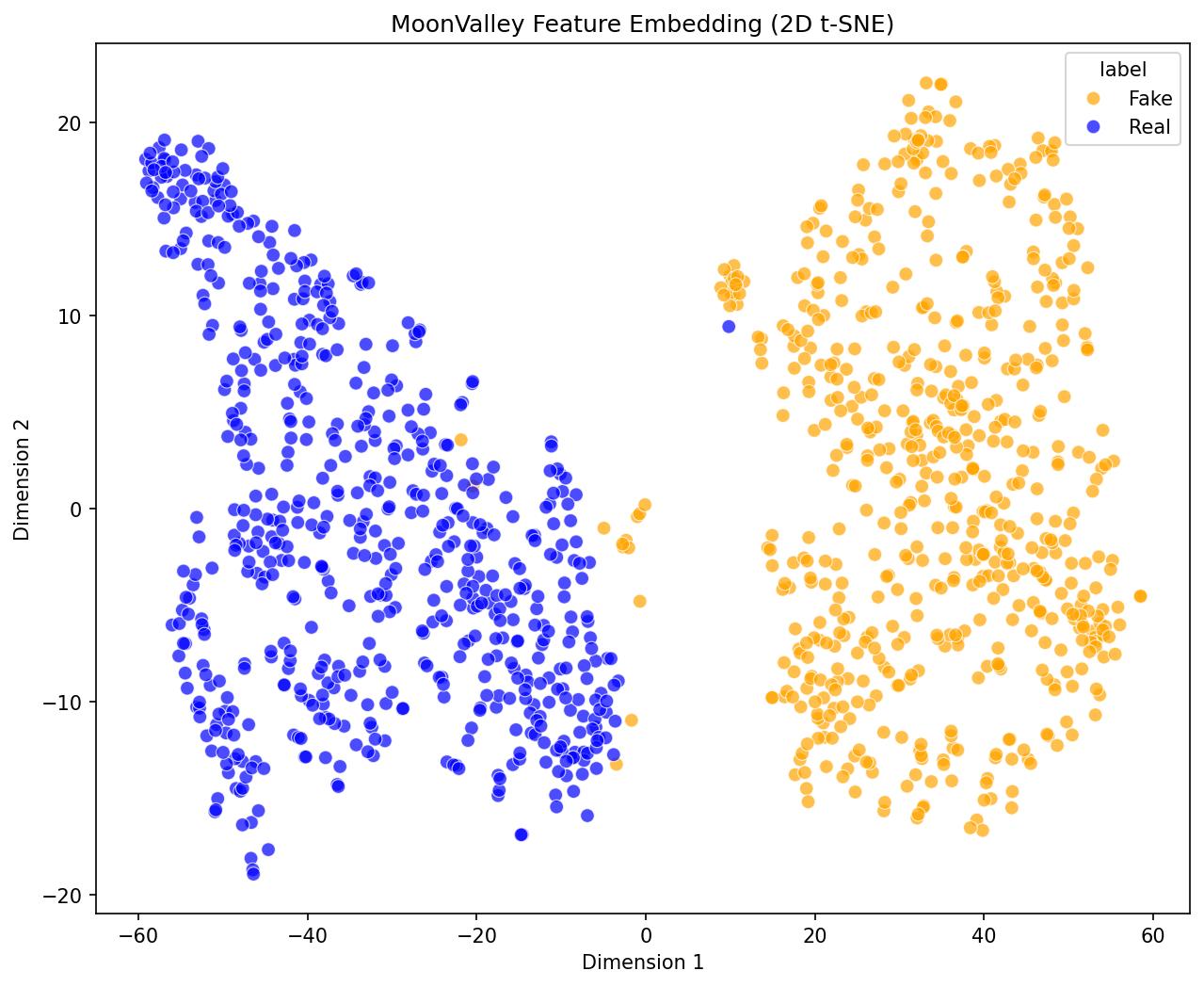}}\hfill
    \subfloat[MorphStudio]{\includegraphics[width=0.19\textwidth]{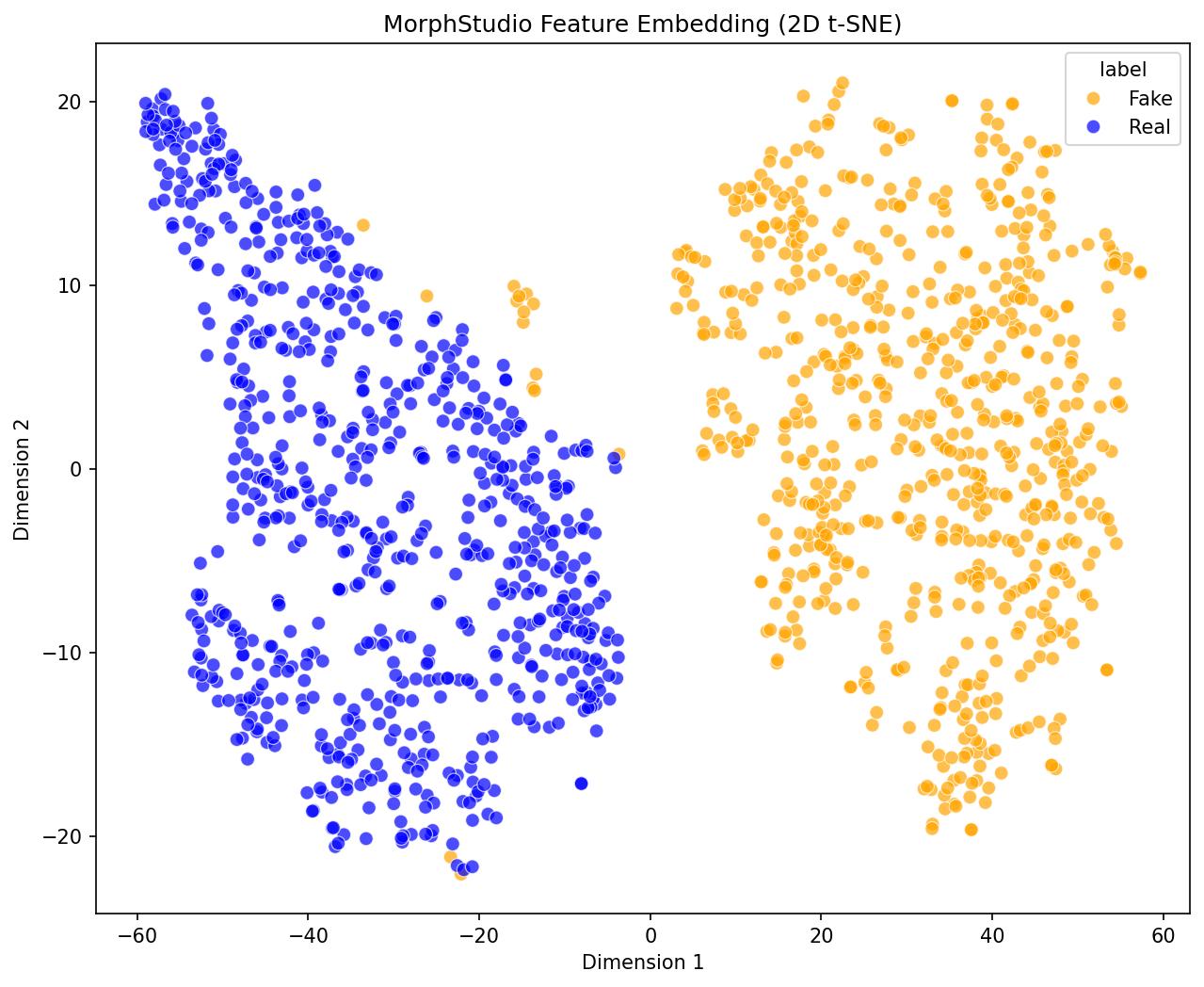}}\hfill
    \subfloat[Show-1]{\includegraphics[width=0.19\textwidth]{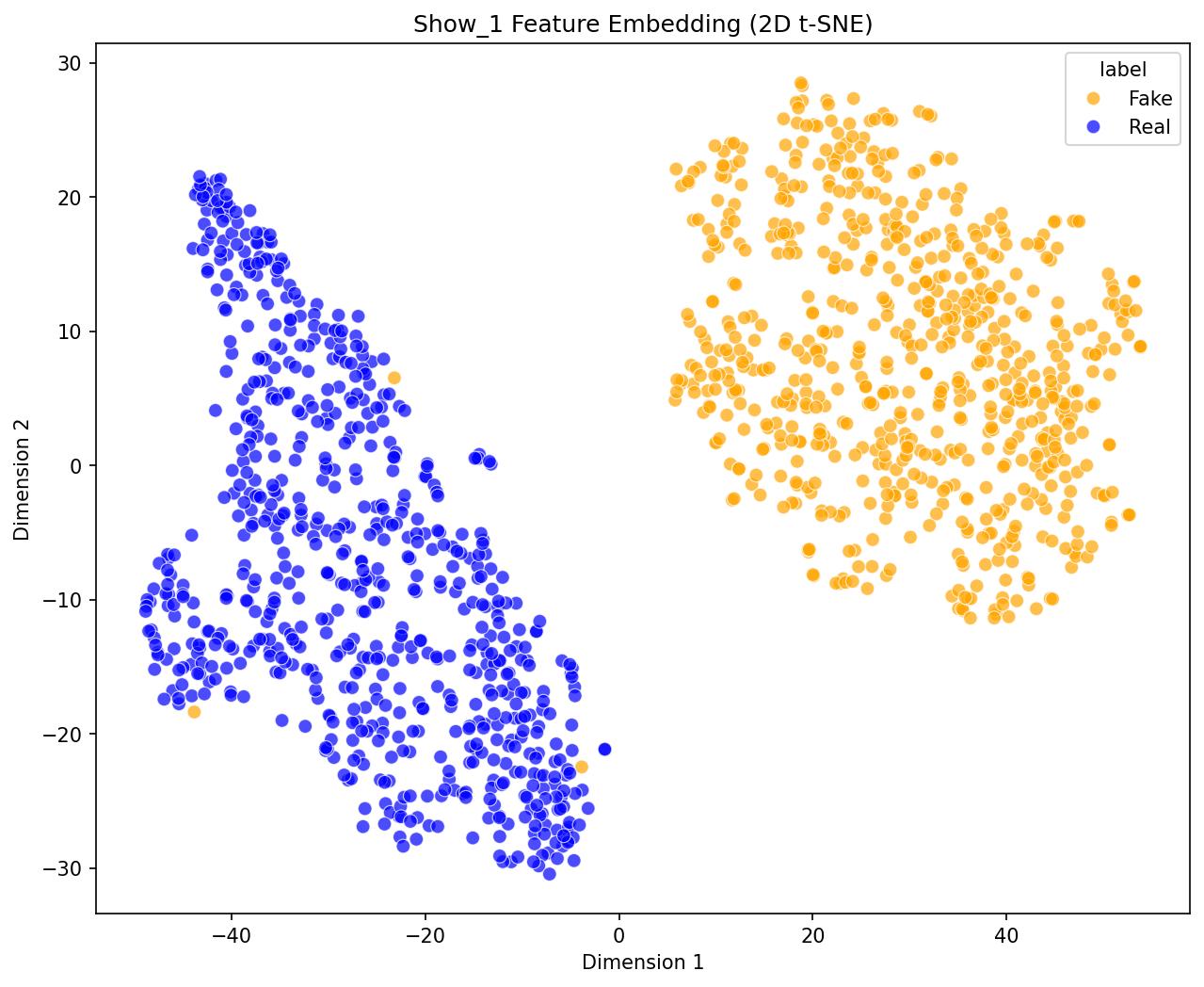}}\hfill
    \subfloat[Sora]{\includegraphics[width=0.19\textwidth]{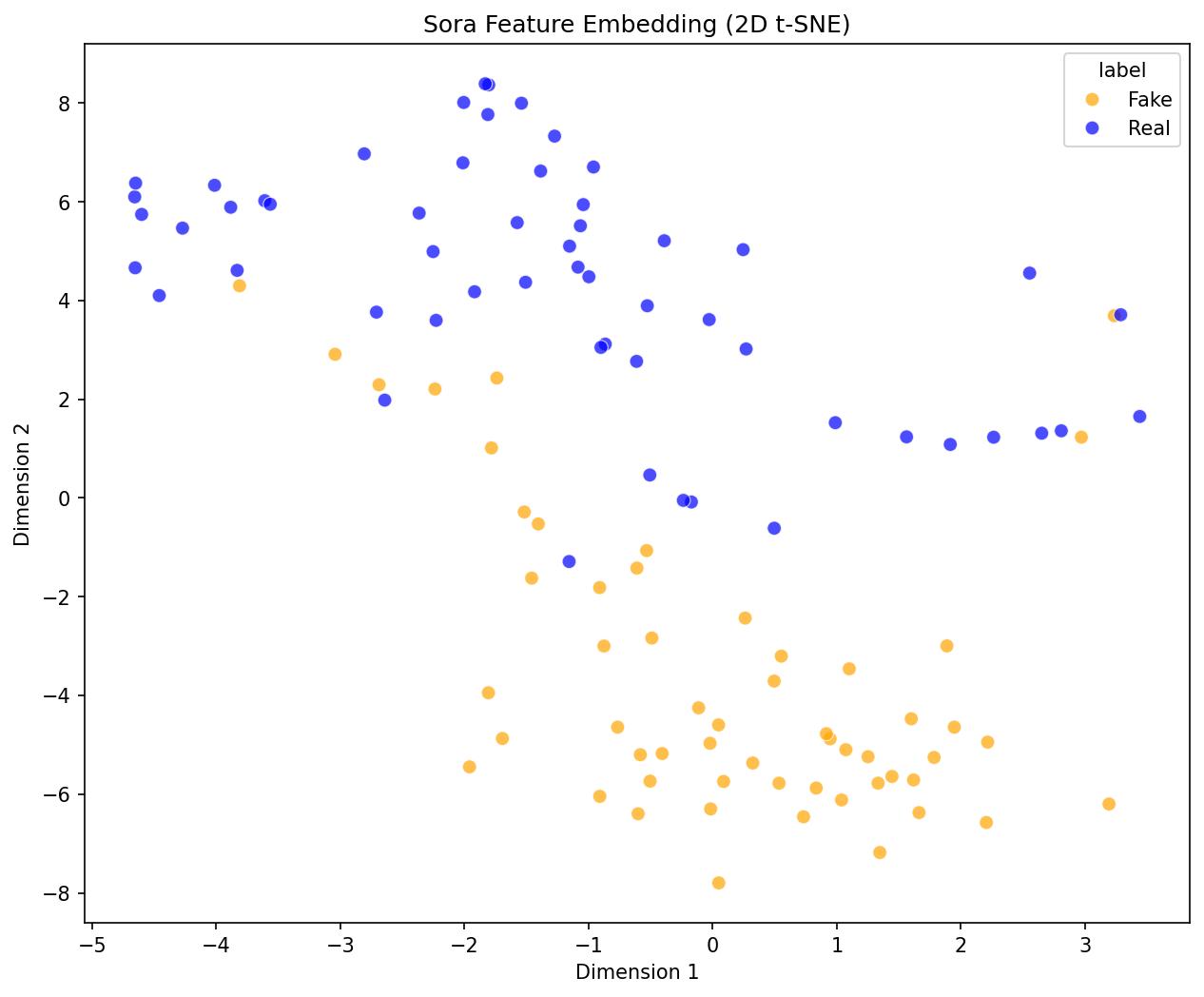}}\hfill
    \subfloat[WildScrape]{\includegraphics[width=0.19\textwidth]{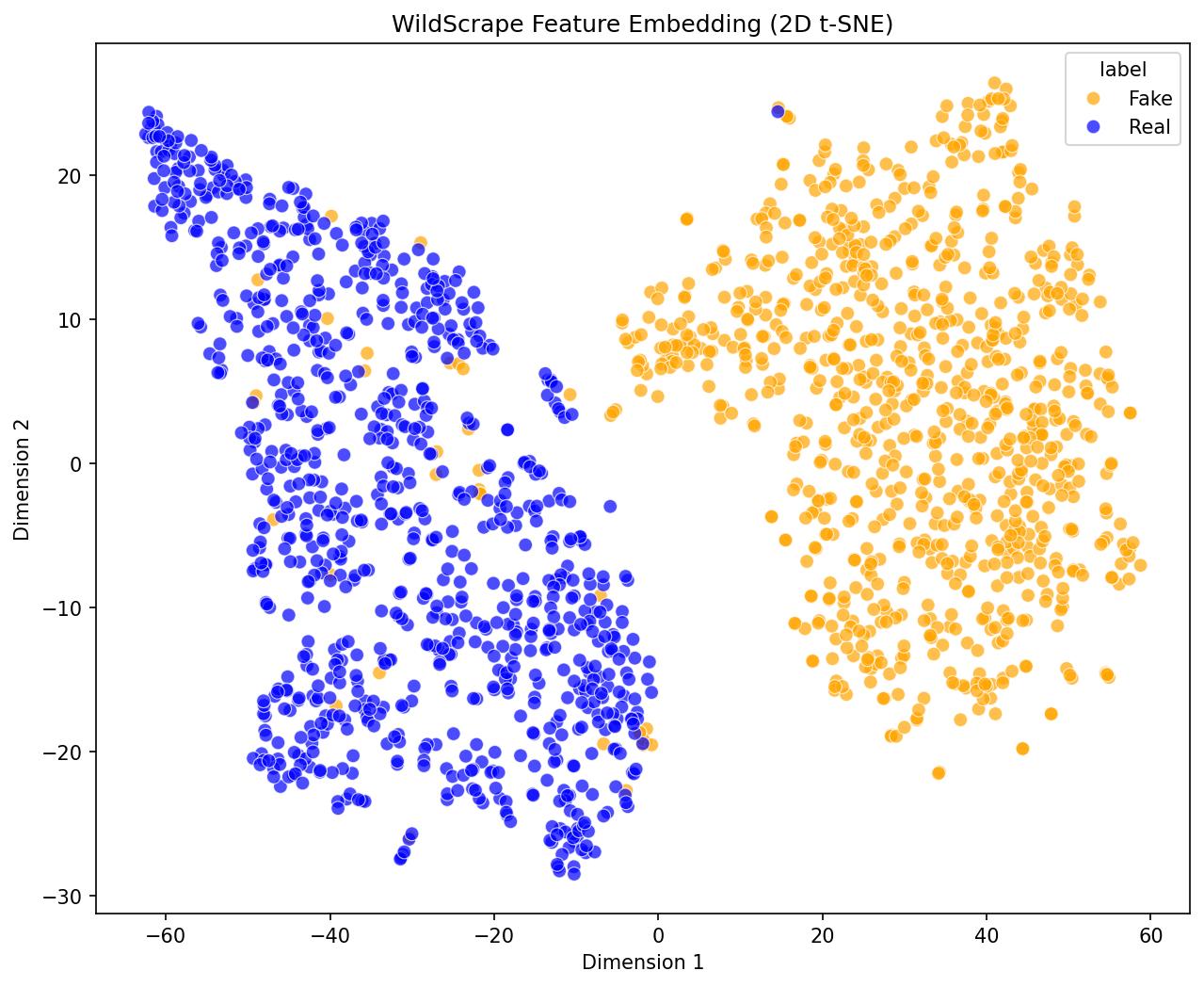}}
    \caption{t-SNE visualizations of 10 subsets on the GenVideo dataset: (a) Crafter, (b) Gen2, (c) HotShot, (d) Lavie, (e) ModelScope, (f) MoonValley, (g) MorphStudio, (h) Show-1, (i) Sora, and (j) WildScrape.}
    \label{fig:tsne_visualization}
\end{figure*}

\begin{figure*}[!ht]
\centering
\includegraphics[width=\textwidth]{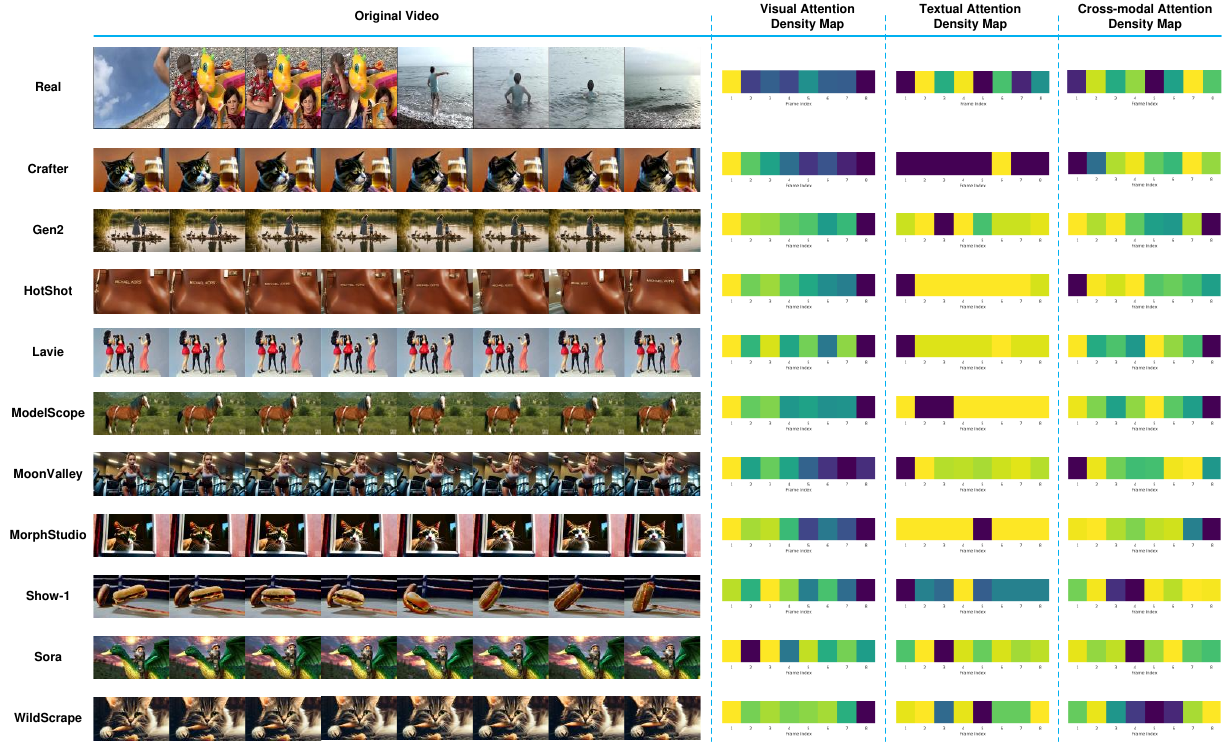} 
\caption{Visualization of Attention Density Maps. Each row displays the attention weights for the visual, textual, and cross-modal branches of a real video and ten AI-generated samples, which are randomly selected from MSR-VTT and each generator subset of GenVideo, respectively.}
\label{fig:density_maps}
\end{figure*}

\subsubsection{Effect of Captioning Backbones and Feature Encoders}
To evaluate the structural flexibility and modular robustness of ATSS, we systematically investigate the impact of various image captioning backbones and visual-textual encoders. 
In our setup, we evaluate the framework by substituting the image caption model with BLIP variants, specifically utilizing the ``blip-image-captioning-base'' and ``blip-image-captioning-large'' weights. 
For visual and textual encoders, we employ the CLIP model configured with ``ViT-B/16'' and ``ViT-B/32'' weights, as well as the X-CLIP model loaded with ``xclip-base-patch16'' and ``xclip-base-patch32'' weights.
The experimental results, summarized in Tables~\ref{tab:ablation2_ap}–\ref{tab:ablation2_acc}, reveal that the synergy between semantic granularity and representation capacity are essential to achieve superior performance.

\noindent \textbf{Impact of Visual-Textual Encoders.}
We evaluate the influence of different visual-textual encoders under fixed image captioning backbones.
As shown in Tables~\ref{tab:ablation2_ap}–\ref{tab:ablation2_acc}, the ``blip2-itm-vit-g'' configuration, which leverages the massive representation capacity of a frozen ViT-g and the explicit Image-Text Matching (ITM) objective, serves as the most effective variant. 
In terms of AP and AUC metrics, this optimal configuration achieves improvements of 0.29\% and 0.12\%, respectively, over the second-best variant utilizing ``xclip-base-patch32''. 
Regarding the ACC metric, ATSS delivers a 2.1\% gain compared with the second-best configuration employing ``ViT-B/16''.
This performance gain is primarily driven by the Image-Text Matching (ITM) head of ``blip2-itm-vit-g'', which significantly enhances cross-modal representation capabilities to expose the subtle, structured generative artifacts inherent in AI-generated videos. 
In contrast, smaller-scale encoders like ``ViT-B/16'' struggle to capture nuanced modal correlations, resulting in a limited ability to distinguish synthetic patterns from authentic dynamics.

\noindent \textbf{Impact of Captioning Backbone.}
We further investigate the role of semantic granularity by evaluating different image captioning backbones under fixed visual-textual encoders. 
As summarized in Tables~\ref{tab:ablation2_ap}–\ref{tab:ablation2_acc}, sophisticated captioning models, particularly ``blip2-flan-t5-xl'', generally provide the most discriminative semantic cues. 
By leveraging advanced captioning models, the approach generates high-quality, fine-grained descriptions that effectively clarify the deterministic trajectories within generative videos, leading to optimal results of 97.56\% AP, 97.11\% AUC and 94.32\% ACC.

However, our results reveal that the benefit of such high-quality captions is highly dependent on the cross-modal representation capability of the visual-textual encoder.
A notable performance inversion is observed where the simpler ``blip-image-captioning-base'' yielding 96.97\% AP outperforms the more complex ``blip2-flan-t5-xl'' which achieves 96.07\% AP when paired with the coarse-grained ``ViT-B/32'' encoder.
This suggests that while detailed textual cues can enhance detection, they also introduce a risk of semantic interference.  
Specifically, coarse-grained textual encoders lack the necessary granularity to resolve fine-grained details when paired with detailed descriptions.
This discrepancy leads to modality alignment bias, where redundant textual information obscures the discriminative patterns within the similarity matrices. 
Conversely, the concise descriptions from ``blip-image-captioning-base'' align more robustly with global visual features, providing a more stable basis for cross-modal matching in configurations with coarse feature granularity (e.g., ``ViT-B/32'').

\subsection{Visualization Analysis}
Fig.~\ref{fig:tsne_visualization} and Fig.~\ref{fig:density_maps} present the visualization results of ATSS on the GenVideo dataset. 
In particular, Fig.~\ref{fig:tsne_visualization} reports the t-SNE projection results, while Fig.~\ref{fig:density_maps} visualizes the attention maps of the three Transformer encoders on a set of cases.

\subsubsection{t-SNE Visualization}
To qualitatively evaluate the discriminative power of the proposed ATSS, we perform 2D t-SNE visualizations on the final fused representations. 
For each generator subset from the GenVideo dataset, we pair the AI-generated videos with an equal number of real videos randomly sampled from MSR-VTT. 
As illustrated in Fig.~\ref{fig:tsne_visualization}, the feature distributions of real and synthetic videos exhibit distinct and robust separation across all ten subsets. 
In particular, even for highly realistic video generators like Sora (Fig.~\ref{fig:tsne_visualization}(i)), where local artifacts are minimized, our model still delineates a clear decision boundary for the majority of samples. 
These results demonstrate that by explicitly modeling temporal self-similarity and anchor-driven evolution, ATSS successfully captures the intrinsic generative fingerprints and achieves reliable cross-generator AIGV detection.

\subsubsection{Attention Density Maps}
To validate the complementarity among the triple-similarity branches and explore the temporal dynamics learned by our model, we visualize a set of attention density maps in Fig.~\ref{fig:density_maps}. 
Specifically, we randomly sample real videos from the MSR-VTT dataset and AI-generated videos produced by each generator from the GenVideo dataset. 
These density maps are then produced by averaging the multi-head self-attention scores along the temporal dimension from each dedicated Transformer encoder, which reveals the frame-level importance emphasized by ATSS. 
The visualization demonstrates that the attention weights fluctuate significantly across different frames and modalities, providing qualitative insights into how the model distinguishes between natural physical continuity and generative coherence.

As shown in Fig.~\ref{fig:density_maps}, real videos exhibit stochastic and heterogeneous attention distributions that reflect the spontaneous and unpredictable nature of authentic scenes.
In contrast, AI-generated videos across all subsets, particularly within the textual and cross-modal branches, display highly structured and sustained attention patterns.
This phenomenon is manifested as the ``blocky'' correlations observed across various generators, which confirms that ATSS successfully captures the unnatural temporal self-similarity and the underlying anchor-driven evolution. 
Furthermore, the distinct divergence in attention peaks across the three streams for the same frame suggests that each branch identifies non-redundant generative fingerprints. 
This inter-modal complementarity ensures that our cross-attentive fusion mechanism effectively integrates multi-faceted clues for robust AIGV detection.

\section{Conclusion}
\label{sec:conclu}
In this work, we introduced ATSS, a unified framework for detecting AI-generated videos by explicitly modeling temporal self-similarity across visual, textual, and cross-modal representations. 
By constructing modality-specific and cross-modal inter-frame similarity matrices and encoding them with dedicated Transformer encoders, ATSS effectively captured fine-grained spatiotemporal and cross-modal artifacts that were typically overlooked by prior approaches. 
A bidirectional cross-attentive fusion mechanism was further proposed to facilitate cross-modal feature interaction, enabling the model to leverage complementary cues at both semantic and visual levels to identify sophisticated generative artifacts. 
Extensive experiments on four challenging benchmarks, including GenVideo, EvalCrafter, VideoPhy, and VidProM, validated the effectiveness of ATSS, which yielded substantial gains in AP, AUC, and ACC over all baseline methods. 
These results demonstrated that ATSS not only achieved superior detection performance but also maintained remarkable stability and adaptability across diverse video generators.

\section*{Acknowledgments}
This research was partially supported by the National Natural Science Foundation of China (62441238, U24B20185).

\bibliographystyle{IEEEtran}
\bibliography{ref}

@InProceedings{Zheng_2025_ICCV,
    author    = {Zheng, Chende and Suo, Ruiqi and Lin, Chenhao and Zhao, Zhengyu and Yang, Le and Liu, Shuai and Yang, Minghui and Wang, Cong and Shen, Chao},
    title     = {D3: Training-Free AI-Generated Video Detection Using Second-Order Features},
    booktitle = {Proceedings of the IEEE/CVF International Conference on Computer Vision (ICCV)},
    month     = {October},
    year      = {2025},
    pages     = {12852-12862}
}

@inproceedings{zhang2025NSGVD,
  title={Physics-Driven Spatiotemporal Modeling for AI-Generated Video Detection},
  author={Zhang, Shuhai and Lian, Zihao and Yang, Jiahao and Li, Daiyuan and Pang, Guoxuan and Liu, Feng and Han, Bo and Li, Shutao and Tan, Mingkui},
  booktitle={Advances in Neural Information Processing Systems},
  year={2025}
}

@inproceedings{
interno2025aigenerated,
title={{AI}-Generated Video Detection via Perceptual Straightening},
author={Christian Intern{\`o} and Robert Geirhos and Markus Olhofer and Sunny Liu and Barbara Hammer and David Klindt},
booktitle={The Thirty-ninth Annual Conference on Neural Information Processing Systems},
year={2025},
url={https://openreview.net/forum?id=LsmUgStXby}
}

@article{
wang2025survey,
title={Survey of Video Diffusion Models: Foundations, Implementations, and Applications},
author={Yimu Wang and Xuye Liu and Wei Pang and Li Ma and Shuai Yuan and Paul Debevec and Ning Yu},
journal={Transactions on Machine Learning Research},
issn={2835-8856},
year={2025},
url={https://openreview.net/forum?id=2ODDBObKjH},
note={Survey Certification}
}

@article{xing2024survey,
  title={A survey on video diffusion models},
  author={Xing, Zhen and Feng, Qijun and Chen, Haoran and Dai, Qi and Hu, Han and Xu, Hang and Wu, Zuxuan and Jiang, Yu-Gang},
  journal={ACM Computing Surveys},
  volume={57},
  number={2},
  pages={1--42},
  year={2024},
  publisher={ACM New York, NY}
}

@InProceedings{Kim_2025_ICCV,
    author    = {Kim, Taehoon and Choi, Jongwook and Jeong, Yonghyun and Noh, Haeun and Yoo, Jaejun and Baek, Seungryul and Choi, Jongwon},
    title     = {Beyond Spatial Frequency: Pixel-wise Temporal Frequency-based Deepfake Video Detection},
    booktitle = {Proceedings of the IEEE/CVF International Conference on Computer Vision (ICCV)},
    month     = {October},
    year      = {2025},
    pages     = {11198-11207}
}

@article{verdoliva2020media,
  title={Media forensics and deepfakes: an overview},
  author={Verdoliva, Luisa},
  journal={IEEE journal of selected topics in signal processing},
  volume={14},
  number={5},
  pages={910--932},
  year={2020},
  publisher={IEEE}
}

@misc{runwayGen42025,
  author       = {{Runway AI, Inc.}},
  title        = {Introducing Runway Gen-4},
  year         = {2025},
  howpublished = {\url{https://runwayml.com/research/introducing-runway-gen-4}},
  note         = {Accessed: 2025-11-10}
}

@article{ji2024distinguish,
  title={Distinguish any fake videos: Unleashing the power of large-scale data and motion features},
  author={Ji, Lichuan and Lin, Yingqi and Huang, Zhenhua and Han, Yan and Xu, Xiaogang and Wu, Jiafei and Wang, Chong and Liu, Zhe},
  journal={arXiv preprint arXiv:2405.15343},
  year={2024}
}

@inproceedings{chen2024videocrafter2,
  title={Videocrafter2: Overcoming data limitations for high-quality video diffusion models},
  author={Chen, Haoxin and Zhang, Yong and Cun, Xiaodong and Xia, Menghan and Wang, Xintao and Weng, Chao and Shan, Ying},
  booktitle={Proceedings of the IEEE/CVF Conference on Computer Vision and Pattern Recognition},
  pages={7310--7320},
  year={2024}
}

@inproceedings{xing2024dynamicrafter,
  title={Dynamicrafter: Animating open-domain images with video diffusion priors},
  author={Xing, Jinbo and Xia, Menghan and Zhang, Yong and Chen, Haoxin and Yu, Wangbo and Liu, Hanyuan and Liu, Gongye and Wang, Xintao and Shan, Ying and Wong, Tien-Tsin},
  booktitle={European Conference on Computer Vision},
  pages={399--417},
  year={2024},
  organization={Springer}
}

@inproceedings{shi2024motion,
  title={Motion-i2v: Consistent and controllable image-to-video generation with explicit motion modeling},
  author={Shi, Xiaoyu and Huang, Zhaoyang and Wang, Fu-Yun and Bian, Weikang and Li, Dasong and Zhang, Yi and Zhang, Manyuan and Cheung, Ka Chun and See, Simon and Qin, Hongwei and others},
  booktitle={ACM SIGGRAPH 2024 Conference Papers},
  pages={1--11},
  year={2024}
}

@inproceedings{li2025image,
  title={Image conductor: Precision control for interactive video synthesis},
  author={Li, Yaowei and Wang, Xintao and Zhang, Zhaoyang and Wang, Zhouxia and Yuan, Ziyang and Xie, Liangbin and Shan, Ying and Zou, Yuexian},
  booktitle={Proceedings of the AAAI Conference on Artificial Intelligence},
  volume={39},
  number={5},
  pages={5031--5038},
  year={2025}
}

@article{barrett2023identifying,
  title={Identifying and mitigating the security risks of generative ai},
  author={Barrett, Clark and Boyd, Brad and Bursztein, Elie and Carlini, Nicholas and Chen, Brad and Choi, Jihye and Chowdhury, Amrita Roy and Christodorescu, Mihai and Datta, Anupam and Feizi, Soheil and others},
  journal={Foundations and Trends{\textregistered} in Privacy and Security},
  volume={6},
  number={1},
  pages={1--52},
  year={2023},
  publisher={Emerald Publishing Limited Boston—Delft}
}

@article{sharma2023survey,
  title={A survey of detection and mitigation for fake images on social media platforms},
  author={Sharma, Dilip Kumar and Singh, Bhuvanesh and Agarwal, Saurabh and Garg, Lalit and Kim, Cheonshik and Jung, Ki-Hyun},
  journal={Applied Sciences},
  volume={13},
  number={19},
  pages={10980},
  year={2023},
  publisher={MDPI}
}

@inproceedings{bao2018towards,
  title={Towards open-set identity preserving face synthesis},
  author={Bao, Jianmin and Chen, Dong and Wen, Fang and Li, Houqiang and Hua, Gang},
  booktitle={Proceedings of the IEEE conference on computer vision and pattern recognition},
  pages={6713--6722},
  year={2018}
}

@article{guo2024pulid,
  title={Pulid: Pure and lightning id customization via contrastive alignment},
  author={Guo, Zinan and Wu, Yanze and Zhuowei, Chen and Zhang, Peng and He, Qian and others},
  journal={Advances in neural information processing systems},
  volume={37},
  pages={36777--36804},
  year={2024}
}

@article{zheng2023out,
  title={Out-of-distribution detection learning with unreliable out-of-distribution sources},
  author={Zheng, Haotian and Wang, Qizhou and Fang, Zhen and Xia, Xiaobo and Liu, Feng and Liu, Tongliang and Han, Bo},
  journal={Advances in neural information processing systems},
  volume={36},
  pages={72110--72123},
  year={2023}
}

@inproceedings{ijcai2021p157,
  title     = {HifiFace: 3D Shape and Semantic Prior Guided High Fidelity Face Swapping},
  author    = {Wang, Yuhan and Chen, Xu and Zhu, Junwei and Chu, Wenqing and Tai, Ying and Wang, Chengjie and Li, Jilin and Wu, Yongjian and Huang, Feiyue and Ji, Rongrong},
  booktitle = {Proceedings of the Thirtieth International Joint Conference on
               Artificial Intelligence, {IJCAI-21}},
  publisher = {International Joint Conferences on Artificial Intelligence Organization},
  editor    = {Zhi-Hua Zhou},
  pages     = {1136--1142},
  year      = {2021},
  month     = {8},
  note      = {Main Track},
  doi       = {10.24963/ijcai.2021/157},
  url       = {https://doi.org/10.24963/ijcai.2021/157},
}

@inproceedings{zhao2023diffswap,
  title={Diffswap: High-fidelity and controllable face swapping via 3d-aware masked diffusion},
  author={Zhao, Wenliang and Rao, Yongming and Shi, Weikang and Liu, Zuyan and Zhou, Jie and Lu, Jiwen},
  booktitle={Proceedings of the IEEE/CVF Conference on Computer Vision and Pattern Recognition},
  pages={8568--8577},
  year={2023}
}

@inproceedings{oorloff2024avff,
  title={Avff: Audio-visual feature fusion for video deepfake detection},
  author={Oorloff, Trevine and Koppisetti, Surya and Bonettini, Nicol{\`o} and Solanki, Divyaraj and Colman, Ben and Yacoob, Yaser and Shahriyari, Ali and Bharaj, Gaurav},
  booktitle={Proceedings of the IEEE/CVF Conference on Computer Vision and Pattern Recognition},
  pages={27102--27112},
  year={2024}
}

@article{li2023learning,
  title={Learning defense transformations for counterattacking adversarial examples},
  author={Li, Jincheng and Zhang, Shuhai and Cao, Jiezhang and Tan, Mingkui},
  journal={Neural Networks},
  volume={164},
  pages={177--185},
  year={2023},
  publisher={Elsevier}
}

@article{blattmann2023stable,
  title={Stable video diffusion: Scaling latent video diffusion models to large datasets},
  author={Blattmann, Andreas and Dockhorn, Tim and Kulal, Sumith and Mendelevitch, Daniel and Kilian, Maciej and Lorenz, Dominik and Levi, Yam and English, Zion and Voleti, Vikram and Letts, Adam and others},
  journal={arXiv preprint arXiv:2311.15127},
  year={2023}
}

@article{zhang2025show,
  title={Show-1: Marrying pixel and latent diffusion models for text-to-video generation},
  author={Zhang, David Junhao and Wu, Jay Zhangjie and Liu, Jia-Wei and Zhao, Rui and Ran, Lingmin and Gu, Yuchao and Gao, Difei and Shou, Mike Zheng},
  journal={International Journal of Computer Vision},
  volume={133},
  number={4},
  pages={1879--1893},
  year={2025},
  publisher={Springer}
}

@inproceedings{NEURIPS2024_6dddcff5,
 author = {Zheng, Chende and Lin, Chenhao and Zhao, Zhengyu and Wang, Hang and Guo, Xu and Liu, Shuai and Shen, Chao},
 booktitle = {Advances in Neural Information Processing Systems},
 editor = {A. Globerson and L. Mackey and D. Belgrave and A. Fan and U. Paquet and J. Tomczak and C. Zhang},
 pages = {59570--59596},
 publisher = {Curran Associates, Inc.},
 title = {Breaking Semantic Artifacts for Generalized AI-generated Image Detection},
 volume = {37},
 year = {2024}
}

@InProceedings{Tan_2024_CVPR,
    author    = {Tan, Chuangchuang and Zhao, Yao and Wei, Shikui and Gu, Guanghua and Liu, Ping and Wei, Yunchao},
    title     = {Rethinking the Up-Sampling Operations in CNN-based Generative Network for Generalizable Deepfake Detection},
    booktitle = {Proceedings of the IEEE/CVF Conference on Computer Vision and Pattern Recognition (CVPR)},
    month     = {June},
    year      = {2024},
    pages     = {28130-28139}
}

@inproceedings{10.1145/3474085.3475508,
author = {Gu, Zhihao and Chen, Yang and Yao, Taiping and Ding, Shouhong and Li, Jilin and Huang, Feiyue and Ma, Lizhuang},
title = {Spatiotemporal Inconsistency Learning for DeepFake Video Detection},
year = {2021},
booktitle = {Proceedings of the 29th ACM International Conference on Multimedia},
pages = {3473–3481},
numpages = {9},
location = {Virtual Event, China},
series = {MM '21}
}

@ARTICLE{10547206,
  author={Coccomini, Davide Alessandro and Zilos, Giorgos Kordopatis and Amato, Giuseppe and Caldelli, Roberto and Falchi, Fabrizio and Papadopoulos, Symeon and Gennaro, Claudio},
  journal={IEEE Transactions on Information Forensics and Security}, 
  title={MINTIME: Multi-Identity Size-Invariant Video Deepfake Detection}, 
  year={2024},
  volume={19},
  number={},
  pages={6084-6096},
  doi={10.1109/TIFS.2024.3409054}
}

@inproceedings{zheng2021exploring,
  title={Exploring Temporal Coherence for More General Video Face Forgery Detection},
  author={Zheng, Yinglin and Bao, Jianmin and Chen, Dong and Zeng, Ming and Wen, Fang},
  booktitle={Proceedings of the IEEE/CVF International Conference on Computer Vision},
  pages={15044--15054},
  year={2021}
}

@inproceedings{xu2023tall,
  title={TALL: Thumbnail Layout for Deepfake Video Detection},
  author={Xu, Yuting and Liang, Jian and Jia, Gengyun and Yang, Ziming and Zhang, Yanhao and He, Ran},
  booktitle={Proceedings of the IEEE/CVF International Conference on Computer Vision},
  pages={22658--22668},
  year={2023}
}

@inproceedings{ni2022expanding,
  title={Expanding language-image pretrained models for general video recognition},
  author={Ni, Bolin and Peng, Houwen and Chen, Minghao and Zhang, Songyang and Meng, Gaofeng and Fu, Jianlong and Xiang, Shiming and Ling, Haibin},
  booktitle={European conference on computer vision},
  pages={1--18},
  year={2022},
  organization={Springer}
}

@inproceedings{bai2024ai,
  title={Ai-generated video detection via spatial-temporal anomaly learning},
  author={Bai, Jianfa and Lin, Man and Cao, Gang and Lou, Zijie},
  booktitle={Chinese Conference on Pattern Recognition and Computer Vision (PRCV)},
  pages={460--470},
  year={2024},
  organization={Springer}
}

@article{chen2024demamba,
  title={Demamba: Ai-generated video detection on million-scale genvideo benchmark},
  author={Chen, Haoxing and Hong, Yan and Huang, Zizheng and Xu, Zhuoer and Gu, Zhangxuan and Li, Yaohui and Lan, Jun and Zhu, Huijia and Zhang, Jianfu and Wang, Weiqiang and others},
  journal={arXiv preprint arXiv:2405.19707},
  year={2024}
}

@inproceedings{li2023blip,
  title={Blip-2: Bootstrapping language-image pre-training with frozen image encoders and large language models},
  author={Li, Junnan and Li, Dongxu and Savarese, Silvio and Hoi, Steven},
  booktitle={International conference on machine learning},
  pages={19730--19742},
  year={2023},
  organization={PMLR}
}

@article{xu2023youku,
  title={Youku-mplug: A 10 million large-scale chinese video-language dataset for pre-training and benchmarks},
  author={Xu, Haiyang and Ye, Qinghao and Wu, Xuan and Yan, Ming and Miao, Yuan and Ye, Jiabo and Xu, Guohai and Hu, Anwen and Shi, Yaya and Xu, Guangwei and others},
  journal={arXiv preprint arXiv:2306.04362},
  year={2023}
}

@inproceedings{xu2016msr,
  title={Msr-vtt: A large video description dataset for bridging video and language},
  author={Xu, Jun and Mei, Tao and Yao, Ting and Rui, Yong},
  booktitle={Proceedings of the IEEE conference on computer vision and pattern recognition},
  pages={5288--5296},
  year={2016}
}

@misc{pika2022,
  author       = {{Pika}},
  title        = {Pika.art},
  year         = {2022},
  howpublished = {\url{https://pika.art/}},
}

@article{wang2023modelscope,
  title={Modelscope text-to-video technical report},
  author={Wang, Jiuniu and Yuan, Hangjie and Chen, Dayou and Zhang, Yingya and Wang, Xiang and Zhang, Shiwei},
  journal={arXiv preprint arXiv:2308.06571},
  year={2023}
}

@misc{morph2023,
  author       = {{Morph Studio}},
  title        = {Morph Studio},
  year         = {2023},
  howpublished = {\url{https://www.morphstudio.com/}}
}

@misc{moonvalley2022,
  author       = {{moonvalley.ai}},
  title        = {moonvalley.ai},
  year         = {2022},
  howpublished = {\url{https://moonvalley.ai/}}
}

@misc{hotshot2023,
  author       = {{Hotshot}},
  title        = {Hotshot-XL},
  year         = {2023},
  howpublished = {\url{https://huggingface.co/hotshotco/Hotshot-XL}}
}

@article{zhang2024show,
  title={Show-1: Marrying pixel and latent diffusion models for text-to-video generation},
  author={Zhang, David Junhao and Wu, Jay Zhangjie and Liu, Jia-Wei and Zhao, Rui and Ran, Lingmin and Gu, Yuchao and Gao, Difei and Shou, Mike Zheng},
  journal={International Journal of Computer Vision},
  pages={1--15},
  year={2024},
  publisher={Springer}
}

@inproceedings{esser2023structure,
  title={Structure and content-guided video synthesis with diffusion models},
  author={Esser, Patrick and Chiu, Johnathan and Atighehchian, Parmida and Granskog, Jonathan and Germanidis, Anastasis},
  booktitle={Proceedings of the IEEE/CVF international conference on computer vision},
  pages={7346--7356},
  year={2023}
}

@article{chen2023videocrafter1,
  title={Videocrafter1: Open diffusion models for high-quality video generation},
  author={Chen, Haoxin and Xia, Menghan and He, Yingqing and Zhang, Yong and Cun, Xiaodong and Yang, Shaoshu and Xing, Jinbo and Liu, Yaofang and Chen, Qifeng and Wang, Xintao and others},
  journal={arXiv preprint arXiv:2310.19512},
  year={2023}
}

@article{wang2025lavie,
  title={Lavie: High-quality video generation with cascaded latent diffusion models},
  author={Wang, Yaohui and Chen, Xinyuan and Ma, Xin and Zhou, Shangchen and Huang, Ziqi and Wang, Yi and Yang, Ceyuan and He, Yinan and Yu, Jiashuo and Yang, Peiqing and others},
  journal={International Journal of Computer Vision},
  volume={133},
  number={5},
  pages={3059--3078},
  year={2025},
  publisher={Springer}
}

@article{brooks2024video,
  title={Video generation models as world simulators},
  author={Brooks, Tim and Peebles, Bill and Holmes, Connor and DePue, Will and Guo, Yufei and Jing, Li and Schnurr, David and Taylor, Joe and Luhman, Troy and Luhman, Eric and others},
  journal={OpenAI Blog},
  volume={1},
  number={8},
  pages={1},
  year={2024}
}

@inproceedings{liu2024evalcrafter,
  title={Evalcrafter: Benchmarking and evaluating large video generation models},
  author={Liu, Yaofang and Cun, Xiaodong and Liu, Xuebo and Wang, Xintao and Zhang, Yong and Chen, Haoxin and Liu, Yang and Zeng, Tieyong and Chan, Raymond and Shan, Ying},
  booktitle={Proceedings of the IEEE/CVF Conference on Computer Vision and Pattern Recognition},
  pages={22139--22149},
  year={2024}
}

@article{bansal2024videophy,
  title={Videophy: Evaluating physical commonsense for video generation},
  author={Bansal, Hritik and Lin, Zongyu and Xie, Tianyi and Zong, Zeshun and Yarom, Michal and Bitton, Yonatan and Jiang, Chenfanfu and Sun, Yizhou and Chang, Kai-Wei and Grover, Aditya},
  journal={arXiv preprint arXiv:2406.03520},
  year={2024}
}

@article{wang2024vidprom,
  title={VidProM: A Million-scale Real Prompt-Gallery Dataset for Text-to-Video Diffusion Models},
  author={Wang, Wenhao and Yang, Yi},
  booktitle={Thirty-eighth Conference on Neural Information Processing Systems},
  year={2024},
}

@inproceedings{wei2024dreamvideo,
  title={Dreamvideo: Composing your dream videos with customized subject and motion},
  author={Wei, Yujie and Zhang, Shiwei and Qing, Zhiwu and Yuan, Hangjie and Liu, Zhiheng and Liu, Yu and Zhang, Yingya and Zhou, Jingren and Shan, Hongming},
  booktitle={Proceedings of the IEEE/CVF Conference on Computer Vision and Pattern Recognition},
  pages={6537--6549},
  year={2024}
}

@article{feng2023dreamoving,
  title={Dreamoving: A human video generation framework based on diffusion models},
  author={Feng, Mengyang and Liu, Jinlin and Yu, Kai and Yao, Yuan and Hui, Zheng and Guo, Xiefan and Lin, Xianhui and Xue, Haolan and Shi, Chen and Li, Xiaowen and others},
  journal={arXiv preprint arXiv:2312.05107},
  year={2023}
}

@inproceedings{xu2024magicanimate,
  title={Magicanimate: Temporally consistent human image animation using diffusion model},
  author={Xu, Zhongcong and Zhang, Jianfeng and Liew, Jun Hao and Yan, Hanshu and Liu, Jia-Wei and Zhang, Chenxu and Feng, Jiashi and Shou, Mike Zheng},
  booktitle={Proceedings of the IEEE/CVF Conference on Computer Vision and Pattern Recognition},
  pages={1481--1490},
  year={2024}
}

@inproceedings{szegedy2016rethinking,
  title={Rethinking the inception architecture for computer vision},
  author={Szegedy, Christian and Vanhoucke, Vincent and Ioffe, Sergey and Shlens, Jon and Wojna, Zbigniew},
  booktitle={Proceedings of the IEEE conference on computer vision and pattern recognition},
  pages={2818--2826},
  year={2016}
}

@article{vaswani2017attention,
  title={Attention is all you need},
  author={Vaswani, Ashish and Shazeer, Noam and Parmar, Niki and Uszkoreit, Jakob and Jones, Llion and Gomez, Aidan N and Kaiser, {\L}ukasz and Polosukhin, Illia},
  journal={Advances in neural information processing systems},
  volume={30},
  year={2017}
}

@inproceedings{nguyen2025vulnerability,
  title={Vulnerability-Aware Spatio-Temporal Learning for Generalizable Deepfake Video Detection},
  author={Nguyen, Dat and Astrid, Marcella and Kacem, Anis and Ghorbel, Enjie and Aouada, Djamila},
  booktitle={Proceedings of the IEEE/CVF International Conference on Computer Vision},
  pages={10786--10796},
  year={2025}
}

@inproceedings{choi2024exploiting,
  title={Exploiting style latent flows for generalizing deepfake video detection},
  author={Choi, Jongwook and Kim, Taehoon and Jeong, Yonghyun and Baek, Seungryul and Choi, Jongwon},
  booktitle={Proceedings of the IEEE/CVF conference on computer vision and pattern recognition},
  pages={1133--1143},
  year={2024}
}

@inproceedings{kundu2025towards,
  title={Towards a universal synthetic video detector: From face or background manipulations to fully ai-generated content},
  author={Kundu, Rohit and Xiong, Hao and Mohanty, Vishal and Balachandran, Athula and Roy-Chowdhury, Amit K},
  booktitle={Proceedings of the Computer Vision and Pattern Recognition Conference},
  pages={28050--28060},
  year={2025}
}

@inproceedings{yan2025generalizing,
  title={Generalizing deepfake video detection with plug-and-play: Video-level blending and spatiotemporal adapter tuning},
  author={Yan, Zhiyuan and Zhao, Yandan and Chen, Shen and Guo, Mingyi and Fu, Xinghe and Yao, Taiping and Ding, Shouhong and Wu, Yunsheng and Yuan, Li},
  booktitle={Proceedings of the Computer Vision and Pattern Recognition Conference},
  pages={12615--12625},
  year={2025}
}

@inproceedings{zhang2024learning,
  title={Learning natural consistency representation for face forgery video detection},
  author={Zhang, Daichi and Xiao, Zihao and Li, Shikun and Lin, Fanzhao and Li, Jianmin and Ge, Shiming},
  booktitle={European Conference on Computer Vision},
  pages={407--424},
  year={2024},
  organization={Springer}
}

@inproceedings{chen2024compressed,
  title={Compressed deepfake video detection based on 3d spatiotemporal trajectories},
  author={Chen, Zongmei and Liao, Xin and Wu, Xiaoshuai and Chen, Yanxiang},
  booktitle={2024 Asia Pacific Signal and Information Processing Association Annual Summit and Conference (APSIPA ASC)},
  pages={1--8},
  year={2024},
  organization={IEEE}
}

@article{pang2023mre,
  title={MRE-Net: Multi-rate excitation network for deepfake video detection},
  author={Pang, Guilin and Zhang, Baopeng and Teng, Zhu and Qi, Zige and Fan, Jianping},
  journal={IEEE Transactions on Circuits and Systems for Video Technology},
  volume={33},
  number={8},
  pages={3663--3676},
  year={2023},
  publisher={IEEE}
}

@inproceedings{feng2023self,
  title={Self-supervised video forensics by audio-visual anomaly detection},
  author={Feng, Chao and Chen, Ziyang and Owens, Andrew},
  booktitle={proceedings of the IEEE/CVF conference on computer vision and pattern recognition},
  pages={10491--10503},
  year={2023}
}

@inproceedings{wang2023altfreezing,
  title={Altfreezing for more general video face forgery detection},
  author={Wang, Zhendong and Bao, Jianmin and Zhou, Wengang and Wang, Weilun and Li, Houqiang},
  booktitle={Proceedings of the IEEE/CVF conference on computer vision and pattern recognition},
  pages={4129--4138},
  year={2023}
}

@inproceedings{cozzolino2021id,
  title={Id-reveal: Identity-aware deepfake video detection},
  author={Cozzolino, Davide and R{\"o}ssler, Andreas and Thies, Justus and Nie{\ss}ner, Matthias and Verdoliva, Luisa},
  booktitle={Proceedings of the IEEE/CVF international conference on computer vision},
  pages={15108--15117},
  year={2021}
}

@inproceedings{chen2022deepfake,
  title={Deepfake detection with spatio-temporal consistency and attention},
  author={Chen, Yunzhuo and Akhtar, Naveed and Haldar, Nur Al Hasan and Mian, Ajmal},
  booktitle={2022 International Conference on Digital Image Computing: Techniques and Applications (DICTA)},
  pages={1--8},
  year={2022},
  organization={IEEE}
}

@article{zhao2023istvt,
  title={ISTVT: interpretable spatial-temporal video transformer for deepfake detection},
  author={Zhao, Cairong and Wang, Chutian and Hu, Guosheng and Chen, Haonan and Liu, Chun and Tang, Jinhui},
  journal={IEEE Transactions on Information Forensics and Security},
  volume={18},
  pages={1335--1348},
  year={2023},
  publisher={IEEE}
}

@article{nie2024dip,
  title={DIP: diffusion learning of inconsistency pattern for general deepfake detection},
  author={Nie, Fan and Ni, Jiangqun and Zhang, Jian and Zhang, Bin and Zhang, Weizhe},
  journal={IEEE Transactions on Multimedia},
  year={2024},
  publisher={IEEE}
}

@article{xie2024shaking,
  title={Shaking the Fake: Detecting Deepfake Videos in Real Time via Active Probes},
  author={Xie, Zhixin and Luo, Jun},
  journal={arXiv preprint arXiv:2409.10889},
  year={2024}
}

@article{liu2024turns,
  title={Turns Out I'm Not Real: Towards Robust Detection of AI-Generated Videos},
  author={Liu, Qingyuan and Shi, Pengyuan and Tsai, Yun-Yun and Mao, Chengzhi and Yang, Junfeng},
  journal={arXiv preprint arXiv:2406.09601},
  year={2024}
}

@INPROCEEDINGS{11210049,
  author={Ma, Long and Yan, Zhiyuan and Guo, Qinglang and Liao, Yong and Yu, Haiyang and Zhou, Pengyuan},
  booktitle={2025 IEEE International Conference on Multimedia and Expo (ICME)}, 
  title={Detecting AI-Generated Video via Frame Consistency}, 
  year={2025},
  volume={},
  number={},
  pages={1-6},
  doi={10.1109/ICME59968.2025.11210049}
}

@inproceedings{2015-kingma,
  title = {Adam: A Method for Stochastic Optimization.},
  author = {Kingma, Diederik P. and Ba, Jimmy},
  booktitle = {Proceedings of the 3rd International Conference on Learning Representations (ICLR 2015)},
  year = 2015
}

\begin{IEEEbiography}[{\includegraphics[width=1in,height=1.25in,clip,keepaspectratio]{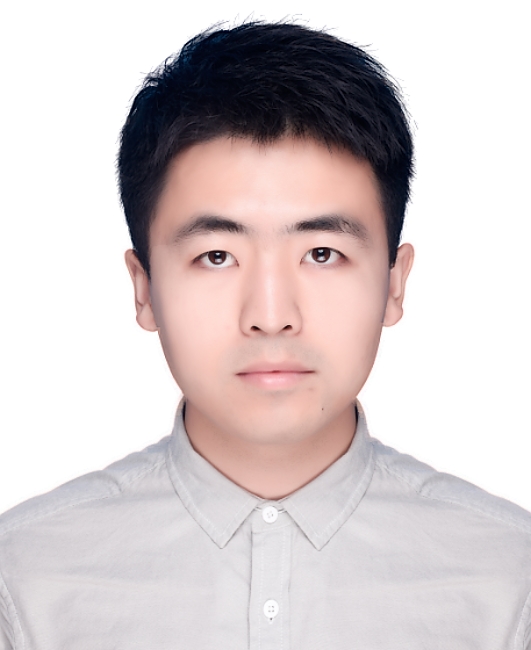}}]{Hang Wang}
received his bachelor's degree in the School of Computer Science and Technology from Xi'an Jiaotong University in 2017, and later earned his Ph.D. from the School of Cyber Science and Engineering at the same university in 2023. He also gained valuable research internship experience at Alibaba DAMO Academy from 2022 to 2023.
Currently, he is an assistant professor at the School of Automation Science and Engineering, Xi'an Jiaotong University. He is also pursuing the second Ph.D. degree with the Department of Computing at The Hong Kong Polytechnic University. His research interests include machine learning, computer vision and image processing.
\end{IEEEbiography}
\vspace{-2mm}

\begin{IEEEbiography}[{\includegraphics[width=1in,height=1.25in,clip,keepaspectratio]{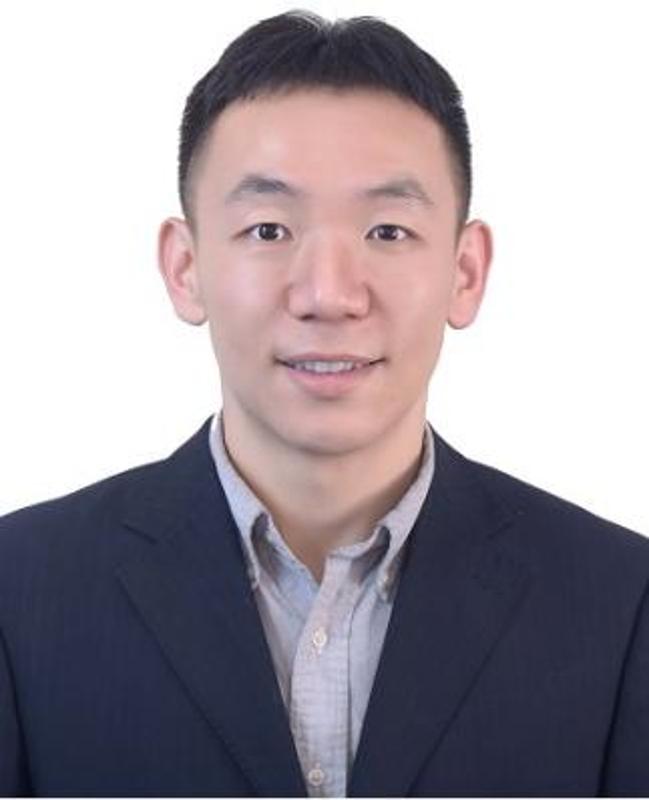}}]{Chao Shen}
(Fellow, IEEE) received the B.S. degree in automation from Xi’an Jiaotong University, China, in 2007, and the Ph.D. degree in control theory and control engineering from Xi’an Jiaotong University, China, in 2014. He is currently a Chair Professor with the School of Cyber Science and Engineering, Xi’an Jiaotong University. His research interests include AI security, insider/intrusion detection, behavioral biometrics, and measurement/experimental methodology.
\end{IEEEbiography}
\vspace{-2mm}

\begin{IEEEbiography}[{\includegraphics[width=1in,height=1.25in,clip,keepaspectratio]{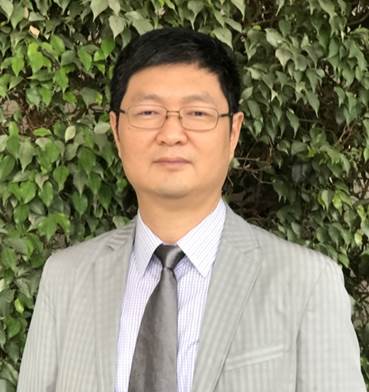}}]{Lei Zhang}
(M’04, SM’14, F’18) received his B.Sc. degree in 1995 from Shenyang Institute of Aeronautical Engineering, Shenyang, P.R. China, and M.Sc. and Ph.D. degrees in Control Theory and Engineering from Northwestern Polytechnical University, Xi’an, P.R. China, in 1998 and 2001, respectively. From 2001 to 2002, he was a research associate in the Department of Computing, The Hong Kong Polytechnic University. From January 2003 to January 2006 he worked as a Postdoctoral Fellow in the Department of Electrical and Computer Engineering, McMaster University, Canada. In 2006, he joined the Department of Computing, The Hong Kong Polytechnic University, as an Assistant Professor. Since July 2017, he has been a Chair Professor in the same department. His research interests include Computer Vision, Image and Video Analysis, Pattern Recognition, and Biometrics, etc. Prof. Zhang has published more than 200 papers in those areas. As of 2026, his publications have been cited more than 120,000 times in literature. Prof. Zhang is a Senior Associate Editor of IEEE Trans. on Image Processing, and is/was an Associate Editor of IEEE Trans. on Pattern Analysis and Machine Intelligence, SIAM Journal of Imaging Sciences, IEEE Trans. on CSVT, and Image and Vision Computing, etc. He is a ``Clarivate Analytics Highly Cited Researcher" from 2015 to 2025. More information can be found in his homepage \url{https://www4.comp.polyu.edu.hk/~cslzhang/}.
\end{IEEEbiography}
\vspace{-2mm}

\begin{IEEEbiography}[{\includegraphics[width=1in,height=1.25in,clip,keepaspectratio]{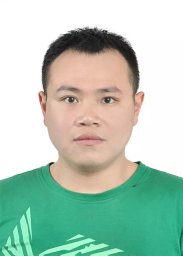}}]{Zhi-Qi Cheng} 
received his B.S. in Computer Science from Southwest Jiaotong University in 2014 and completed his Ph.D. in 2019. During his doctoral studies, he was a joint Ph.D. student at the City University of Hong Kong (2016-2017) and later at Carnegie Mellon University (2017-2019). He also gained valuable experience through internships at Alibaba DAMO Academy (2016), Google Brain (2018), and Microsoft Research (2019). From 2019 to 2022, he served as a postdoctoral research associate at the School of Computer Science of Carnegie Mellon University (CMU). He is currently a Project Scientist at the Language Technologies Institute (LTI), part of the School of Computer Science at CMU. He significantly contributed to several key projects, including DARPA's AIDA, KAIROS, IARPA's DIVA, and NIST's PSIAP. His research has had a substantial impact, such as being utilized in the Washington Post's coverage of the Capitol riots, for which he was awarded the Pulitzer Prize for Public Service. Additionally, he has been honored with the Intel Ph.D. Fellowship and IBM Outstanding Student Scholarship.
\end{IEEEbiography}

\end{document}